\documentclass{article}

\usepackage[nonatbib, final]{neurips_2020}

\usepackage[numbers,sort]{natbib}

\usepackage[utf8]{inputenc} 
\usepackage[T1]{fontenc}    
\usepackage{hyperref}       
\usepackage{url}            
\usepackage{booktabs}       
\usepackage{amsfonts}       
\usepackage{nicefrac}       
\usepackage{microtype}      

\usepackage{graphicx}
\usepackage{color}

\usepackage{booktabs} 

\usepackage{amssymb}
\usepackage{amsmath}
\usepackage{subcaption}

\usepackage{placeins}
\usepackage{wrapfig}
\usepackage{bm}

\usepackage[capitalize]{cleveref}

\title{Woodbury Transformations for \\Deep Generative Flows}

\author{%
  You Lu \\
  Department of Computer Science\\
  Virginia Tech \\
  Blacksburg, VA\\
  \texttt{you.lu@vt.edu} \\
   \And
   Bert Huang \\
   Department of Computer Science \\
   Tufts University \\
   Medford, MA \\
   \texttt{bert@cs.tufts.edu} \\
}

\begin{document}

\maketitle
\begin{abstract}
	Normalizing flows are deep generative models that allow efficient likelihood calculation and sampling. The core requirement for this advantage is that they are constructed using functions that can be efficiently inverted and for which the determinant of the function's Jacobian can be efficiently computed. Researchers have introduced various such flow operations, but few of these allow rich interactions among variables without incurring significant computational costs. In this paper, we introduce \emph{Woodbury transformations}, which achieve efficient invertibility via the Woodbury matrix identity and efficient determinant calculation via Sylvester's determinant identity. In contrast with other operations used in state-of-the-art normalizing flows, Woodbury transformations enable (1) high-dimensional interactions, (2) efficient sampling, and (3) efficient likelihood evaluation. Other similar operations, such as 1x1 convolutions, emerging convolutions, or periodic convolutions allow at most two of these three advantages. In our experiments on multiple image datasets, we find that Woodbury transformations allow learning of higher-likelihood models than other flow architectures while still enjoying their efficiency advantages.
\end{abstract}
\label{sec:abstract}

\section{Introduction}
\label{sec:introduction}

Deep generative models are powerful tools for modeling complex distributions and have been applied to many tasks such as synthetic data generation~\cite{oord2016wavenet,yu2017seqgan}, domain adaption~\cite{zhu2017unpaired}, and structured prediction~\cite{sohn2015learning}. Examples of these models include autoregressive models~\cite{graves2013generating,oord2016pixel},  variational autoencoders~\cite{kingma2013auto,rezende2014stochastic}, generative adversarial networks~\cite{goodfellow2014generative}, and normalizing flows~\cite{dinh2014nice,rezende2015variational,dinh2016density,kingma2018glow}. Normalizing flows are special because of two advantages: They allow efficient and exact computation of log-likelihood and sampling.

Flow-based models are composed of a series of invertible functions, which are specifically designed so that their inverse and determinant of the Jacobian are easy to compute. However, to preserve this computational efficiency, these functions usually cannot sufficiently encode dependencies among dimensions of a variable. For example, affine coupling layers~\cite{dinh2014nice} split a variable to two parts and require the second part to only depend on the first. But they ignore the dependencies among dimensions in the second part.

To address this problem, \citet{dinh2014nice,dinh2016density} introduced a fixed permutation operation that reverses the ordering of the channels of pixel variables. \citet{kingma2018glow} introduced a 1$\times$1 convolution, which are a generalized permutation layer, that uses a weight matrix to model the interactions among dimensions along the channel axis. Their experiments demonstrate the importance of capturing dependencies among dimensions. Relatedly, \citet{hoogeboom2019emerging} proposed emerging convolution operations, and \citet{hoogeboom2019emerging} and \citet{Finzi2019Invertible} proposed periodic convolution. These two convolution layers have $d\times d$ kernels that can model dependencies along the spatial axes in addition to the channel axis. However, the increase in representational power comes at a cost: These convolution operations do not scale well to high-dimensional variables. 
The emerging convolution is a combination of two autoregressive convolutions~\cite{germain2015made,kingma2016improved}, whose inverse is not parallelizable.
To compute the inverse or determinant of the Jacobian, the periodic convolution requires transforming the input and the convolution kernel to Fourier space. This transformation is computationally costly. 

In this paper, we develop \emph{Woodbury transformations} for generative flows. Our method is also a generalized permutation layer and uses spatial and channel transformations to model dependencies among dimensions along spatial and channel axes. We use the Woodbury matrix identity~\cite{woodbury1950inverting} and Sylvester's determinant identity~\cite{sylvester1851} to compute the inverse and Jacobian determinant, respectively, so that both the training and sampling time complexities are linear to the input variable's size. We also develop a memory-efficient variant of the Woodbury transformation, which has the same advantage as the full transformation but uses significantly reduced memory when the variable is high-dimensional. In our experiments, we found that Woodbury transformations enable model quality comparable to many state-of-the-art flow architectures while maintaining significant efficiency advantages. 

\section{Deep Generative Flows}
\label{sec:background}

In this section, we briefly introduce the deep generative flows. More background knowledge can be found in the appendix.

A normalizing flow~\cite{rezende2015variational} is composed of a series of invertible functions $\mathbf{f} = \mathbf{f}_1 \circ \mathbf{f}_2 \circ ... \circ \mathbf{f}_K$, which transform $\mathbf{x}$ to a latent code $\mathbf{z}$ drawn from a simple distribution. Therefore, with the \emph{change of variables} formula, we can rewrite the log-likelihood $\log p_{\theta}(\mathbf{x})$ to be
\begin{equation}
\log p_{\theta}(\mathbf{x}) = \log p_{Z}(\mathbf{z}) + \sum_{i=1}^{K} \log \left|\det \left(\frac{\partial \mathbf{f}_i}{\partial \mathbf{r}_{i-1}}\right)\right|,
\label{eq:relikelihood}
\end{equation}
where $\mathbf{r}_i = \mathbf{f}_i(\mathbf{r}_{i-1})$, $\mathbf{r}_{0} = \mathbf{x}$, and $\mathbf{r}_{K}=\mathbf{z}$. 

Flow-based generative models~\cite{dinh2014nice,dinh2016density,kingma2018glow} are developed on the theory of normalizing flows. Each transformation function used in the models is a specifically designed neural network that has a tractable Jacobian determinant and inverse. We can sample from a trained flow $\mathbf{f}$ by computing $\mathbf{z} \sim p_{Z}(\mathbf{z}), \quad \mathbf{x} = \mathbf{f}^{-1}(\mathbf{z})$.

There have been many operations, i.e., layers, proposed in recent years for generative flows. In this section, we discuss some commonly used ones, and more related works will be discussed in \cref{sec:related}. 

\textbf{Actnorm layers}~\cite{kingma2018glow} perform per-channel affine transformations of the activations using scale and bias parameters to improve training stability and performance. The actnorm is formally expressed as
$\mathbf{y}_{:,i,j} = \mathbf{s} \odot \mathbf{x}_{:,i,j} + \mathbf{b}$,
where both the input $\mathbf{x}$ and the output $\mathbf{y}$ are $c \times h \times w$ tensors, $c$ is the channel dimension, and $h \times w$ are spatial dimensions. The parameters $\mathbf{s}$ and $\mathbf{b}$ are $c \times 1$ vectors. 

\textbf{Affine coupling layers}~\cite{dinh2014nice,dinh2016density} split the input $\mathbf{x}$ into two parts, $\mathbf{x}_a, \mathbf{x}_b$.  And then fix $\mathbf{x}_a$ and force $\mathbf{x}_b$ to only relate to $\mathbf{x}_a$, so that the Jacobian is a triangular matrix. Formally, we compute
\begin{eqnarray*}
	&&\mathbf{x}_a, \mathbf{x}_b = \text{split}(\mathbf{x}), \quad\quad\quad\quad~~ \mathbf{y}_a = \mathbf{x}_a,  \\
	&&\mathbf{y}_b = \mathbf{s}(\mathbf{x}_a) \odot \mathbf{x}_b  + \mathbf{b}(\mathbf{x}_a), \quad \mathbf{y} = \text{concat}(\mathbf{y}_a, \mathbf{y}_b),
\end{eqnarray*}
where $\mathbf{s}$ and $\mathbf{b}$ are two neural networks with $\mathbf{x}_a$ as input. The $\text{split}$ and the $\text{concat}$ split and concatenate the variables along the channel axis. Usually, $s$ is restricted to be positive. An additive coupling layer is a special case when $\mathbf{s}=\mathbf{1}$. 

Actnorm layers only rescale the dimensions of $\mathbf{x}$, and affine coupling layers only relate $\mathbf{x}_b$ to $\mathbf{x}_a$ but omit dependencies among different dimensions of $\mathbf{x}_b$. Thus, we need other layers to capture local dependencies among dimensions.

\textbf{Invertible convolutional layers}~\cite{kingma2018glow,hoogeboom2019emerging,Finzi2019Invertible} are generalized permutation layers that can capture correlations among dimensions.  The 1$\times$1 convolution~\cite{kingma2018glow} is $\mathbf{y}_{:,i,j} = \mathbf{Mx}_{:,i,j}$, where $\mathbf{M}$ is a $c \times c$ matrix. The Jacobian of a 1$\times$1 convolution is a block diagonal matrix, so that its log-determinant is $hw\log |\det(\mathbf{M})|$. Note that the 1$\times$1 convolution only operates along the channel axis and ignores the dependencies along the spatial axes.

\begin{wrapfigure}[18]{r}{0.47\textwidth}
	\begin{minipage}{0.42\linewidth}
		\centering
		\includegraphics[width=\linewidth]{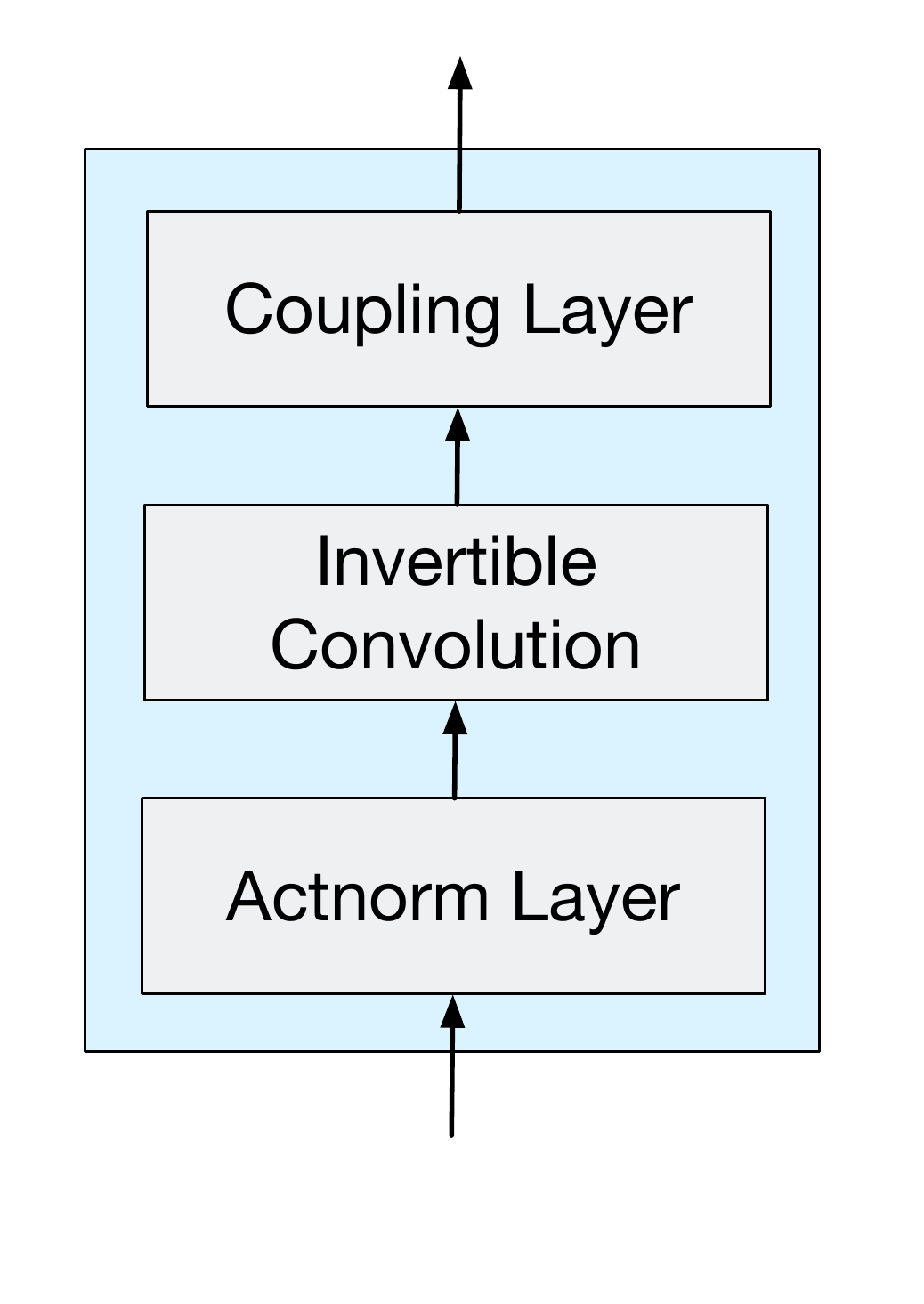}
		\subcaption{Flow step}
		\label{fig:flow-step}
	\end{minipage}
	\hspace{0.25cm}
	\begin{minipage}{0.48\linewidth}
		\centering
		\includegraphics[width=\linewidth]{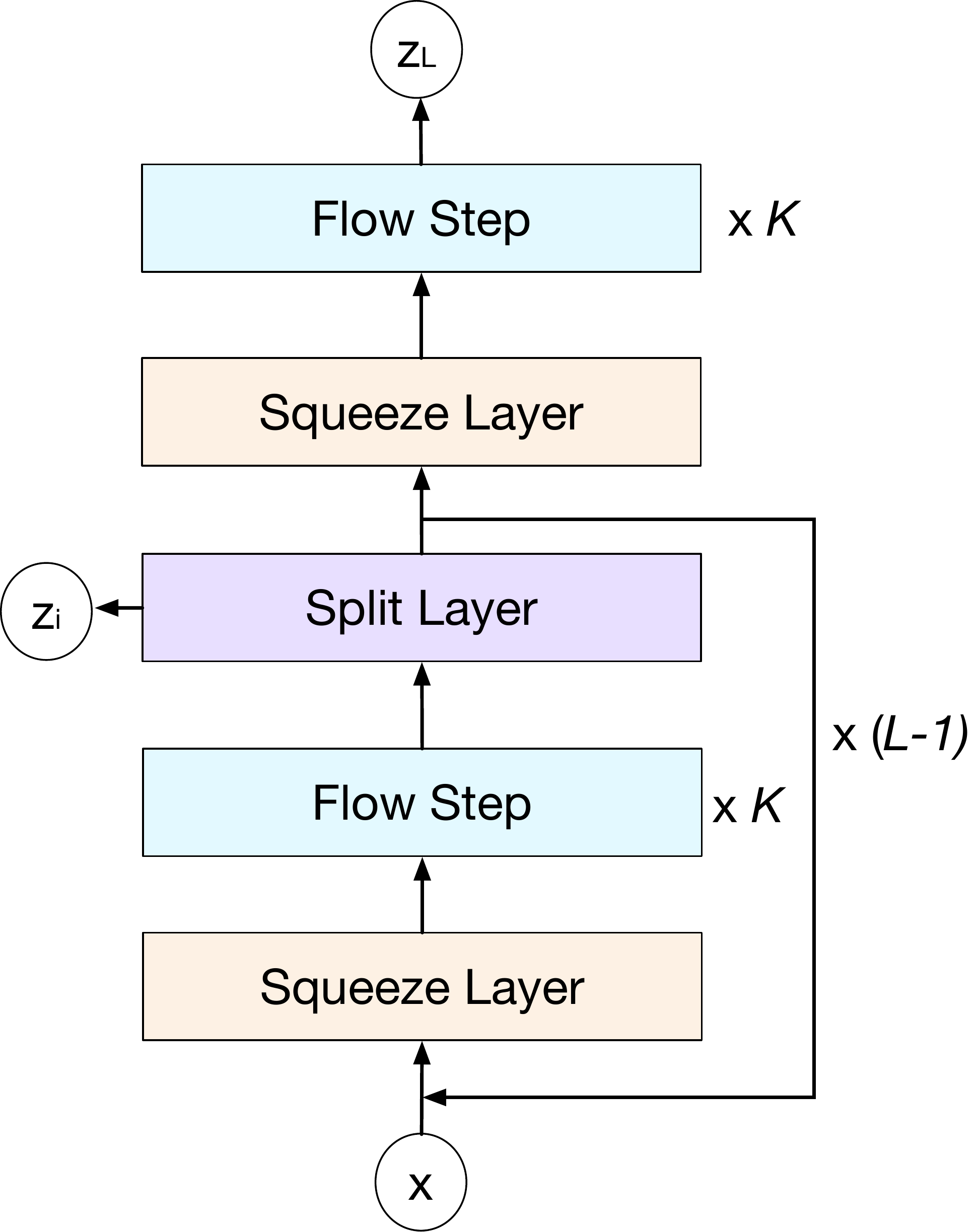}
		\subcaption{Multi-scale arch.}
		\label{fig:multi-scale}
	\end{minipage}
	\vspace{-3pt}
	\caption{Overview of architecture of generative flows. We can design the flow step by selecting a suitable convolutional layer and a coupling layer based on the task. Glow~\cite{kingma2018glow} uses 1$\times$1 convolutions and affine coupling.}
	\label{fig:flow-arc}
\end{wrapfigure}

Emerging convolutions~\cite{hoogeboom2019emerging} combine two autoregressive convolutions~\cite{germain2015made,kingma2016improved}. Each autoregressive convolution masks out some weights to force an autoregressive structure, so that the Jacobian is a triangular matrix and computing its determinant is efficient. One problem of emerging convolution is the computation of inverse is non-parallelizable, so that is inefficent for high-dimensional variables.

Periodic convolutions~\cite{hoogeboom2019emerging,Finzi2019Invertible} transform the input and kernel to the Fourier domain using discrete Fourier transformations, so the convolution function is an element-wise matrix product with a block-diagonal Jacobian. The computational cost of periodic convolutions is $\mathcal{O}(chw\log(hw) +c^3hw)$. Thus, when the input is high-dimensional, both training and sampling are expensive.

\textbf{Multi-scale architectures}~\cite{dinh2016density} compose flow layers to generate rich models, using \emph{split layers} to factor out variables and \emph{squeeze layers} to shuffle dimensions, resulting in an architecture with $K$ flow steps and $L$ levels. See \cref{fig:flow-arc}.

\section{Woodbury Transformations}
\label{sec:method}

In this section, we introduce Woodbury transformations as an efficient means to model high-dimensional correlations.

\subsection{Channel and Spatial Transformations}

Suppose we reshape the input $\mathbf{x}$ to be a $c \times n$ matrix, where $n = hw$. Then the 1$\times$1 convolution can be reinterpreted as a matrix transformation
\begin{equation}
\mathbf{y} = \mathbf{W}^{(c)}\mathbf{x},
\label{eq:c-trans}
\end{equation}
where $\mathbf{y}$ is also a $c \times n$ matrix, and $\mathbf{W}^{(c)}$ is a $c \times c$ matrix. For consistency, we will call this a channel transformation. For each column $\mathbf{x}_{:,i}$, the correlations among channels are modeled by $\mathbf{W}^{(c)}$. However, the correlation between any two rows  $\mathbf{x}_{:,i}$ and $\mathbf{x}_{:,j}$ is not captured. 
Inspired by Eq.~\ref{eq:c-trans}, we use a spatial transformation to model interactions among dimensions along the spatial axis
\begin{equation}
\mathbf{y} = \mathbf{xW}^{(s)},
\label{eq:s-trans}
\end{equation}
where $\mathbf{W}^{(s)}$ is an $n \times n$ matrix that models the correlations of each row $\mathbf{x}_{i,:}$. 
Combining Equation~\ref{eq:c-trans} and Equation~\ref{eq:s-trans}, we have
\begin{eqnarray}
\mathbf{x}_c = \mathbf{W}^{(c)}\mathbf{x}, ~~~~~~~~
\mathbf{y} = \mathbf{x}_c\mathbf{W}^{(s)}.
\label{eq:combin-trans}
\end{eqnarray}

For each dimension of output $\mathbf{y}_{i, j}$, we have
$\mathbf{y}_{i,j} = \sum_{v=1}^{c}\left(\sum_{u=1}^{n}\mathbf{W}^{(c)}_{i,u}\cdot \mathbf{x}_{u,v}\right)\cdot \mathbf{W}^{(s)}_{v,j}$.

Therefore, the spatial and channel transformations together can model the correlation between any pair of dimensions. However, in this preliminary form, directly using Eq.~\ref{eq:combin-trans} is inefficient for large $c$ or $n$. First, we would have to store two large matrices $\mathbf{W}^c$ and $\mathbf{W}^s$, so the space cost is $\mathcal{O}(c^2 + n^2)$. Second, the computational cost of Eq.~\ref{eq:combin-trans} is $\mathcal{O}(c^2n + n^2c)$---quadratic in the input size. Third, the computational cost of the Jacobian determinant is $\mathcal{O}(c^3+n^3)$, which is far too expensive in practice. 

\subsection{Woodbury Transformations}

We solve the three scalability problems by using a low-rank factorization. Specifically, we define
\begin{eqnarray*}
	\mathbf{W}^{(c)} = \mathbf{I}^{(c)} + \mathbf{U}^{(c)}\mathbf{V}^{(c)}, ~~~~~~~~
	\mathbf{W}^{(s)} = \mathbf{I}^{(s)} + \mathbf{U}^{(s)}\mathbf{V}^{(s)},
	\label{eq:W}
\end{eqnarray*}
where $\mathbf{I}^{(c)}$ and $\mathbf{I}^{(s)}$ are $c$- and $n$-dimensional identity matrices, respectively. The matrices $\mathbf{U}^c$, $\mathbf{V}^c$, $\mathbf{U}^s$, and $\mathbf{V}^s$ are of size $c \times d_c$, $d_c \times c$, $n \times d_s$, and $d_c \times n$, respectively, where $d_c$ and $d_s$ are constant latent dimensions of these four matrices. Therefore, we can rewrite Equation~\ref{eq:combin-trans} as
\begin{eqnarray}
\mathbf{x}_c = (\mathbf{I}^{(c)} + \mathbf{U}^{(c)}\mathbf{V}^{(c)}) \mathbf{x}, ~~~~~~~~
\mathbf{y} =  \mathbf{x}_c(\mathbf{I}^{(s)} + \mathbf{U}^{(s)}\mathbf{V}^{(s)}).
\label{eq:w-trans}
\end{eqnarray}

We call Eq.~\ref{eq:w-trans} the Woodbury transformation because the Woodbury matrix identity \cite{woodbury1950inverting} and Sylvester's determinant identity \cite{sylvester1851} allow efficient computation of its inverse and Jacobian determinant.

\textbf{Woodbury matrix identity.}\footnote{A more general version replaces $\mathbf{I}^{(n)}$ and $\mathbf{I}^{(k)}$ with arbitrary invertible $n \times n$ and $k \times k$ matrices. But this simplified version is sufficient for our tasks.} Let $\mathbf{I}^{(n)}$ and $\mathbf{I}^{(k)}$ be $n$- and $k$-dimensional identity matrices, respectively. Let $\mathbf{U}$ and $\mathbf{V}$ be $n \times k$ and $k \times n$ matrices, respectively. If $\mathbf{I}^{(k)}+\mathbf{VU}$ is invertible, then		
$(\mathbf{I}^{(n)} + \mathbf{UV})^{-1} = \mathbf{I}^{(n)} - \mathbf{U}(\mathbf{I}^{k} + \mathbf{VU})^{-1}\mathbf{V}$.

\textbf{Sylvester's determinant identity.} Let $\mathbf{I}^{(n)}$ and $\mathbf{I}^{(k)}$ be $n$- and $k$-dimensional identity matrices, respectively. Let $\mathbf{U}$ and $\mathbf{V}$ be $n \times k$ and $k \times n$ matrices, respectively. Then,
$\det(\mathbf{I}^{(n)} + \mathbf{UV}) = \det(\mathbf{I}^{(k)}+\mathbf{VU})$.

Based on these two identities, we can efficiently compute the inverse and Jacobian determinant
\begin{eqnarray}
\mathbf{x}_c = \mathbf{y}(\mathbf{I}^{(s)} - \mathbf{U}^{(s)}(\mathbf{I}^{(d_s)} + \mathbf{V}^{(s)}\mathbf{U}^{(s)})^{-1}\mathbf{V}^{(s)}), \nonumber\\
\mathbf{x} =   (\mathbf{I}^{(c)} - \mathbf{U}^{(c)}(\mathbf{I}^{(d_c)} + \mathbf{V}^{(c)}\mathbf{U}^{(c)})^{-1}\mathbf{V}^{(c)})\mathbf{x}_c,
\label{eq:w-inverse}
\end{eqnarray}
and
\begin{eqnarray}
\log \left| \det \left(\frac{\partial \mathbf{y}}{\partial \mathbf{x}} \right) \right|  &= n \log\left|\det(\mathbf{I}^{(d_c)}+\mathbf{V}^{(c)}\mathbf{U}^{(c)})\right| 
+ c \log\left|\det(\mathbf{I}^{(d_s)}+\mathbf{V}^{(s)}\mathbf{U}^{(s)})\right|,
\label{eq:w-det}
\end{eqnarray}
where $\mathbf{I}^{(d_c)}$ and $\mathbf{I}^{(d_s)}$ are $d_c$- and $d_s$-dimensional identity matrices, respectively.

\begin{figure}[tp]
	\centering
	\begin{subfigure}[t]{0.3\textwidth}
		\centering
		\includegraphics[width=0.6\linewidth]{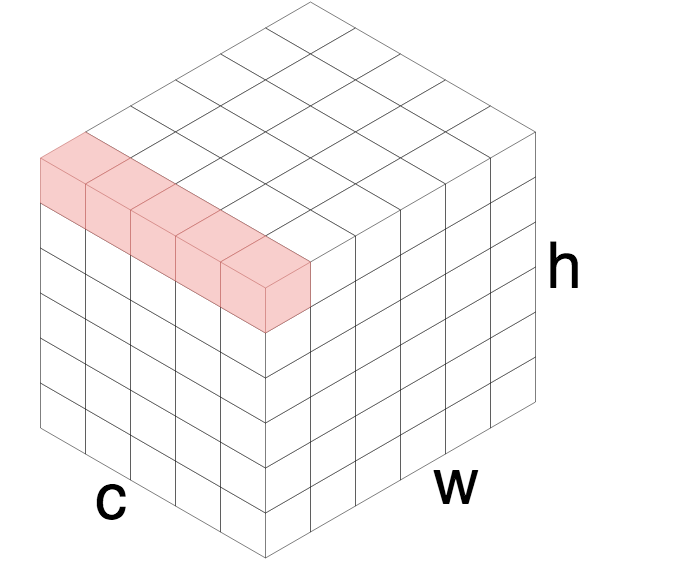}  
		\caption{1$\times$1 convolution}
		\label{fig:conv}
	\end{subfigure}
	\begin{subfigure}[t]{0.3\textwidth}
		\centering
		\includegraphics[width=0.6\linewidth]{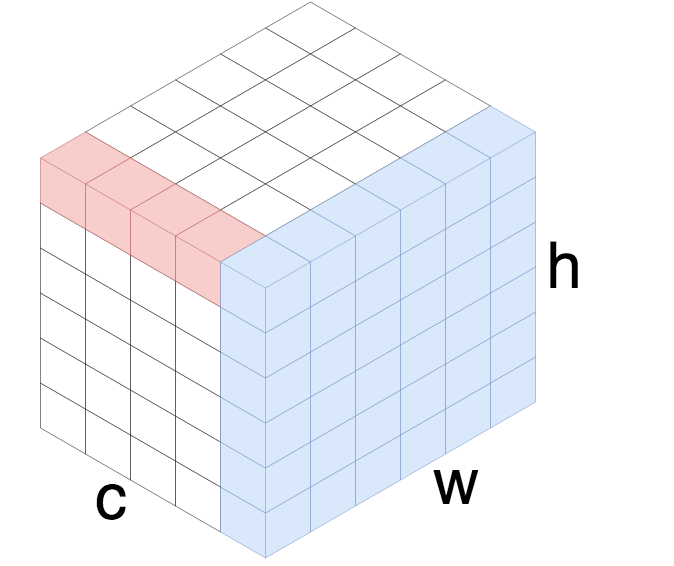}  
		\caption{Woodbury}
		\label{fig:w}
	\end{subfigure}
	\begin{subfigure}[t]{0.3\textwidth}
		\centering
		\includegraphics[width=0.6\linewidth]{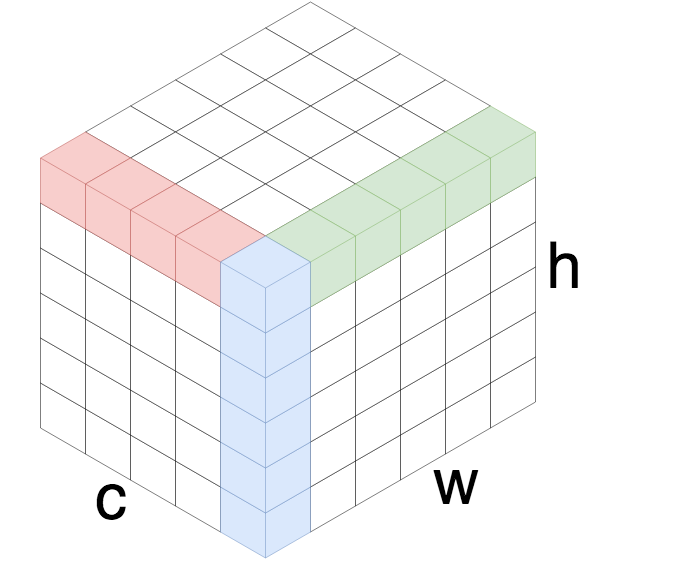}  
		\caption{ME-Woodbury}
		\label{fig:me-w}
	\end{subfigure}
	\caption{Visualization of three transformations. The 1$\times$1 convolution only operates along the channel axis. The Woodbury transformation operates along both the channel and spatial axes, modeling the dependencies of one channel directly via one transformation. The ME-Woodbury transformation operates along three axes. It uses two transformations to model spatial dependencies.}
	\label{fig:three-trans}
\end{figure}

A Woodbury transformation is also a generalized permutation layer. We can directly replace an invertible convolution in Figure~\ref{fig:flow-step} with a Woodbury transformation. In contrast with 1$\times$1 convolutions, Woodbury transformations are able to model correlations along both channel and spatial axes. We illustrate this in Figure~\ref{fig:three-trans}. To implement Woodbury transformations, we need to store four weight matrices, i.e., $\mathbf{U}^{(c)}, \mathbf{U}^{(s)}, \mathbf{V}^{(c)}$, and $\mathbf{V}^{(s)}$. To simplify our analysis, let $d_c \le d$ and $d_s \le d$, where $d$ is a constant. This setting is also consistent with our experiments. The size of $\mathbf{U}^{(c)}$ and $\mathbf{V}^{(c)}$ is $\mathcal{O}(dc)$, and the size of $\mathbf{U}^{(c)}$ and $\mathbf{V}^{(c)}$ is $\mathcal{O}(dn)$. The space complexity is $\mathcal{O}(d(c+n))$.

For training and likelihood computation, the main computational bottleneck is computing $\mathbf{y}$ and the Jacobian determinant.  To compute $\mathbf{y}$ with  Equation~\ref{eq:combin-trans}, we need to first compute the channel transformation and then compute the spatial transformation. The computational complexity is $\mathcal{O}(dcn)$. To compute the determinant with Equation~\ref{eq:w-det}, we need to first compute the matrix product of $\mathbf{V}$ and $\mathbf{U}$, and then compute the determinant. The computational complexity is $\mathcal{O}(d^2(c + n) + d^3)$. 

For sampling, we need to compute the inverse transformations, i.e., Equation~\ref{eq:w-inverse}. With the Woodbury identity, we actually only need to compute the inverses of $\mathbf{I}^{(d_s)} + \mathbf{V}^{(s)}\mathbf{U}^{(s)}$ and $\mathbf{I}^{(d_c)} + \mathbf{V}^{(c)}\mathbf{U}^{(c)}$, which are computed with time complexity $\mathcal{O}(d^3)$. To implement the inverse transformations, we can compute the matrix chain multiplication, so we can avoid computing the product of two large matrices twice, yielding cost $\mathcal{O}(c^2+n^2)$. For example, for the inverse spatial transformation, we can compute it as 
$\mathbf{x}_c = \mathbf{y} - ((\mathbf{yU}^{(s)})(\mathbf{I}^{(d_s)} + \mathbf{V}^{(s)}\mathbf{U}^{(s)})^{-1})\mathbf{V}^{(s)}$,
so that its complexity is $\mathcal{O}(d^3+cd^2+cnd)$. The total computational complexity of Equation~\ref{eq:w-inverse} is $\mathcal{O}(dcn + d^2(n+c)+d^3)$. 

In practice, we found that for a high-dimensional input, a relatively small $d$ is enough to obtain good performance, e.g., the input is $256 \times 256 \times 3$ images, and $d=16$. In this situation, $nc \ge d^3$. Therefore, we can omit $d$ and approximately see the spatial complexity as $\mathcal{O}(c+n)$, and the forward or inverse transformation as $\mathcal{O}(nc)$. They are all linear to the input size.

We do not restrict $\mathbf{U}$ and $\mathbf{V}$ to force $\mathbf{W}$ to be invertible. Based on analysis by \citet{hoogeboom2019emerging}, the training maximizes the log-likelihood, which implicitly pushes $ \det (\mathbf{I} + \mathbf{VU})$ away from $0$. Therefore, it is not necessary to explicitly force invertibility. In our experiments, the Woodbury transformations are as robust as other invertible convolution layers.

\subsection{Memory-Efficient Variant}

In Eq.~\ref{eq:combin-trans}, one potential challenge arises from the sizes of $\mathbf{U}^{(s)}$ and $\mathbf{V}^{((s))}$, which are linear in $n$. The challenge is that $n$ may be large in some practical problems, e.g., high-resolution images. We develop a memory-efficient variant of Woodbury transformations, i.e., ME-Woodbury, to solve this problem. The ME version can effectively reduce space complexity from $\mathcal{O}(d(c+hw))$ to $\mathcal{O}(d(c+h+w))$.

The difference between ME-Woodbury transformations and Woodbury transformations is that the ME form cannot directly model spatial correlations. As shown in Figure~\ref{fig:me-w}, it uses two transformations, for height and width, together to model the spatial correlations. Therefore, for a specific channel $k$, when two dimensions $\mathbf{x}_{k,i,j}$ and $\mathbf{x}_{k,u,v}$ are in two different heights, and widths, their interaction will be modeled indirectly. In our experiments, we found that this limitation only slightly impacts ME-Woodbury's performance. More details on ME-Woodbury transformations are in the appendix.

\section{Related Work}
\label{sec:related}

\citet{rezende2015variational} developed planar flows for variational inference
$\mathbf{z}_{t+1} = \mathbf{z}_{t} + \mathbf{u}\delta(\mathbf{w}^T\mathbf{z}_{t} + b)$,
where $\mathbf{z}$, $\mathbf{w}$, and $\mathbf{u}$ are $d$-dimensional vectors, $\delta()$ is an activation function, and $b$ is a scalar. 

\citet{berg2018sylvester} generalized these to Sylvester flows
$\mathbf{z}_{t+1} = \mathbf{z}_t + \mathbf{QR}\delta(\tilde{\mathbf{R}}\mathbf{Q}^T\mathbf{z}_{t} + \mathbf{r})$,
where $\mathbf{R}$ and $\tilde{\mathbf{R}}$ are upper triangular matrices, $\mathbf{Q}$ is composed of a set of orthonormal vectors, and $\mathbf{r}$ is a $d$-dimensional vector.  
The resulting Jacobian determinant can be efficiently computed via Sylvester's identity, just as our methods do. However, Woodbury transformations have key differences from Sylvester flows. First, \citeauthor{berg2018sylvester} only analyze their models on vectors. The inputs to our layers are matrices, so our method operates on high-dimensional input, e.g., images. Second, though Sylvester flows are inverse functions, computing their inverse is difficult. One possible way is to apply iterative methods~\cite{behrmann2018invertible,song2019mintnet,chen2019residual} to compute the inverse. But this research direction is unexplored. Our layers can be inverted efficiently with the Woodbury identity. Third, our layers do not restrict the transformation matrices to be triangular or orthogonal. In fact, Woodbury transformations can be seen as another generalized variant of planar flows on matrices, with  $\delta(\mathbf{x}) = \mathbf{x}$, and whose inverse is tractable. Roughly speaking, Woodbury transformations can also be viewed as applying the planar flows sequentially to each row of the input matrix. After this work was completed and submitted, we learned that the TensorFlow software~\cite{abadi2016tensorflow} also uses the Woodbury identity in their affine bijector.

Normalizing flows have also been used for variational inference, density estimation, and generative modeling. Autoregressive flows~\cite{kingma2016improved,papamakarios2017masked,huang2018neural,macow2019} restrict each variable to depend on those that precede it in a sequence, forcing a triangular Jacobian. Non-linear coupling layers replace the affine transformation function. Specifically, spline flows~\cite{muller2019neural,durkan2019neural} use spline interpolation, and Flow++~\cite{ho2019flow} uses a mixture cumulative distribution function to define these functions. Flow++ also uses variational dequantization to prevent model collapse. Many works ~\cite{kingma2018glow,hoogeboom2019emerging,Finzi2019Invertible,karami2019invertible} develop invertible convolutional flows to model interactions among dimensions. MintNet~\cite{song2019mintnet} is a flexible architecture composed of multiple masked invertible layers. I-ResNet~\cite{behrmann2018invertible,chen2019residual} uses discriminative deep network architecture as the flow. These two models require iterative methods to compute the inverse. Discrete flows~\cite{tran2019discrete,hoogeboom2019integer} and latent flows~\cite{ziegler2019latent} can be applied to discrete data such as text. Continuous-time flows~\cite{chen2018neural,grathwohl2018ffjord} have been developed based on the theory of ordinary differential equations.

\section{Experiments}
\label{sec:experiments}

In this section, we compare the performance of Woodbury transformations against other modern flow architectures, measuring running time, bit per-dimension ($\log_2$-likelihood), and sample quality.

\begin{wrapfigure}[25]{r}{0.55\textwidth}
	\centering
	\vspace{-10pt}
	\includegraphics[width=1.0\linewidth]{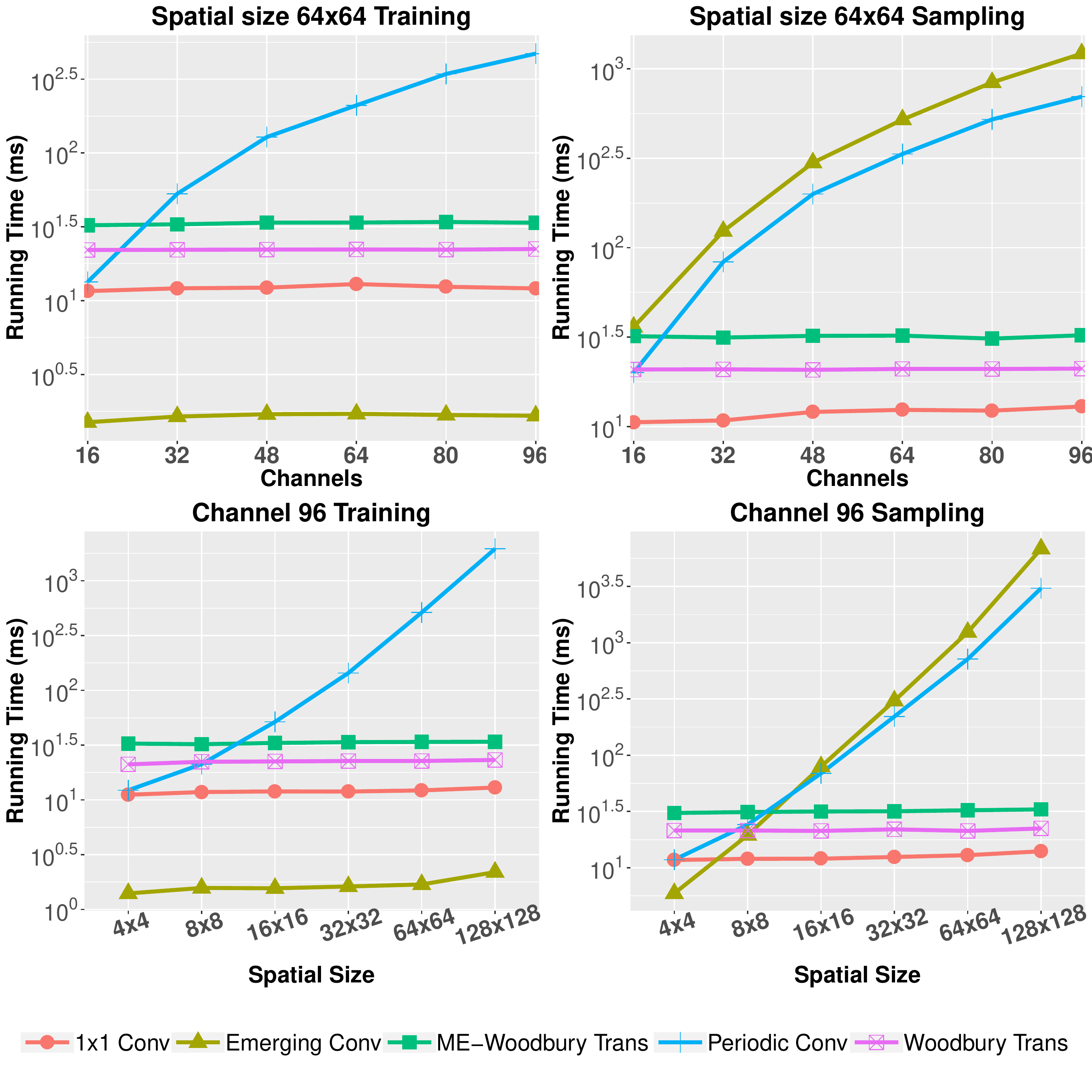}
	\vspace{-17pt}
	\caption{Running time comparison. Sampling with emerging convolutions is slow, since their inverses are not parallelizable. Periodic convolutions are costly for larger inputs. Both 1$\times$1 convolutions and Woodbury transformations are efficient in training and sampling.}
	\label{fig:rt}
\end{wrapfigure}

\textbf{Running Time}~
We follow \citet{Finzi2019Invertible} and compare the per-sample running time of Woodbury transformations to other generalized permutations: 1$\times$1~\cite{kingma2018glow}, emerging~\cite{hoogeboom2019emerging}, and periodic convolutions~\cite{hoogeboom2019emerging,Finzi2019Invertible}. We test the training time and sampling time. In training, we compute (1) forward propagation, i.e., $\mathbf{y}=\mathbf{f}(\mathbf{x})$, of a given function $\mathbf{f}()$, (2) the Jacobian determinant, i.e., $\det\left(\left|\frac{\partial \mathbf{y}}{\partial \mathbf{x}}\right|\right)$, and (3) the gradient of parameters. For sampling, we compute the inverse of transformation $\mathbf{x} = \mathbf{f}^{-1}(\mathbf{y})$.  For emerging and periodic convolutions, we use $3\times 3$ kernels. For Woodbury transformations, we fix the latent dimension $d=16$. For fair comparison, we implement all methods in Pytorch and run them on an Nvidia Titan V GPU. We follow \citet{hoogeboom2019emerging} and implement the emerging convolution inverse in Cython, and we compute it on a 4 Ghz CPU (the GPU version is slower than the Cython version). We first fix the spatial size to be $64 \times 64$ and vary the channel number. We then fix the channel number to be $96$ and vary the spatial size.

The results are shown in Figure~\ref{fig:rt}. For training, the emerging convolution is the fastest. This is because its Jacobian is a triangular matrix, so computing its determinant is much more efficient than other methods. The Woodbury transformation and ME-Woodbury are slightly slower than the 1x1 convolution, since they contain more transformations. Emerging convolutions, Woodbury transformations, and 1x1 convolutions only slightly increase with input size, rather than increasing with $\mathcal{O}(c^3)$. This invariance to input size is likely because of how the GPU parallelizes computation. The periodic convolution is efficient only when the input size is small. When the size is large, it becomes slow, e.g., when the input size is $96\times 64 \times 64$, it is around $30$ times slower than Woodbury transformations. In our experiments, we found that the Fourier transformation requires a large amount of memory. According to \citet{Finzi2019Invertible}, the Fourier step may be the bottleneck that impacts periodic convolution's scalability. A more efficient implementation of Fourier transformation, e.g., \cite{karami2019invertible}, may improve its running time.

For sampling, both 1$\times$1 convolutions and Woodbury transformations are efficient. The 1$\times$1 convolution is the fastest, and the Woodbury transformations are only slightly slower. Neither is sensitive to the change of input size. Emerging convolutions and periodic convolutions are much slower than Woodbury transformations, and their running time increases with the input size. When the input size is $96 \times 128 \times 128$, they are around $100$ to $200$ times slower than Woodbury transformations. This difference is because emerging convolutions cannot make use of parallelization, and periodic transformations require conversion to Fourier form. Based on these results, we can conclude that both emerging convolution and periodic convolution do not scale well to high-dimensional inputs. In contrast, Woodbury transformations are efficient in both training and sampling.

\textbf{Quantitative Evaluation}~
We compare Woodbury transformations with state-of-the-art flow models, measuring bit per-dimension (bpd). We train with the CIFAR-10~\cite{krizhevsky2009learning} and ImageNet~\cite{russakovsky2015imagenet} datasets. We compare with three generalized permutation methods---1$\times$1 convolution, emerging convolution, and periodic convolution---and two coupling layers---neural spline coupling~\cite{durkan2019neural} and MaCow~\cite{macow2019}. We use Glow (Fig.~\ref{fig:flow-arc}, \cite{kingma2018glow}) as the basic flow architecture. For each method, we replace the corresponding layer. For example, to construct a flow with Woodbury transformations, we replace the 1$\times$1 convolution with a Woodbury transformation, i.e., Eq.~\ref{eq:combin-trans}. For all generalized permutation methods, we use affine coupling. For each of the coupling layer baselines, we substitute it for the affine coupling. We tune the parameters of neural spline coupling and MaCow so that their sizes are close to affine coupling. We follow  \citet{hoogeboom2019emerging} and test the performance of small models. For $32\times 32$ images, we set the number of levels to $L=3$ and the number of steps per-level to $K=8$. For $64\times 64$ images, we use $L=4$ and $K=16$. More details are in the appendix.

\begin{table}[htp]
	\begin{center}
		\caption{Quantitative evaluation results.}
		\label{tab:nll}
		\begin{tabular}{ l  l  l  l | l l}
			\toprule
			~ & \multicolumn{3}{c|}{\textbf{Quantitative measure (bpd)}} & \multicolumn{2}{c}{\textbf{Model sizes (\# parameters)}} \\
			\toprule
			~ & CIFAR-10 & ImageNet & ImageNet & 32x32 images & 64x64 images \\
			~ & 32x32 & 32x32 & 64x64 & ~ & ~\\
			\midrule
			1$\times$1 convolution & 3.51 & 4.32 & 3.94 & 11.02M & 37.04M  \\
			Emerging & 3.48 & 4.26 & 3.91 & 11.43M & 40.37M \\
			Periodic & 3.49 & 4.28 & 3.92 & 11.21M&38.61M \\
			Neural spline & 3.50 & 4.24 & 3.95 & \textbf{10.91M} & 38.31M  \\
			MaCow & 3.48 & 4.34 & 4.15 & 11.43M & 37.83M \\
			ME-Woodbury & 3.48 & 4.22 & 3.91 & 11.02M & \textbf{36.98M} \\
			Woodbury & \textbf{3.47} & \textbf{4.20} & \textbf{3.87} & 11.10M & 37.60M\\
			\bottomrule
		\end{tabular}
	\end{center}
\end{table}

\begin{figure*}[!t]
	\centering
	\begin{subfigure}[t]{0.31\textwidth}
		\centering
		\includegraphics[width=0.775\linewidth]{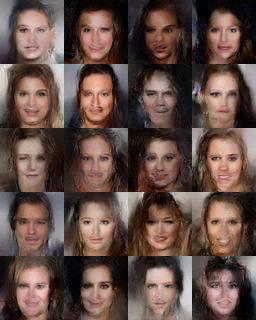}
		\caption*{Woodbury-Glow\\ Iteration 50,000}
	\end{subfigure}
	\begin{subfigure}[t]{0.31\textwidth}
		\centering
		\includegraphics[width=0.775\linewidth]{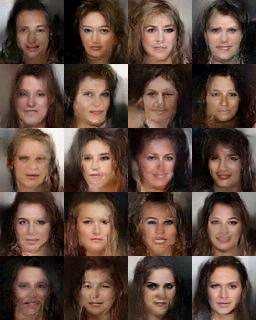}  
		\caption*{Woodbury-Glow \\ Iteration 100,000}
	\end{subfigure}
	\begin{subfigure}[t]{0.31\textwidth}
		\centering
		\includegraphics[width=0.775\linewidth]{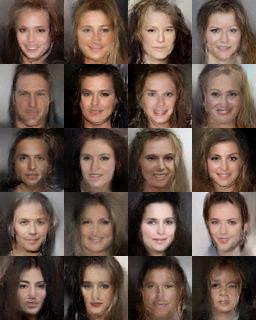}
		\caption*{Woodbury-Glow \\ Iteration 600,000}
	\end{subfigure}
	\begin{subfigure}[t]{0.31\textwidth}
		\centering
		\includegraphics[width=0.775\linewidth]{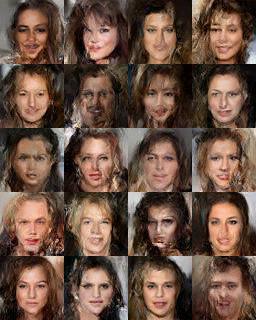}  
		\caption*{Glow \\ Iteration 50,000}
	\end{subfigure}
	\begin{subfigure}[t]{0.31\textwidth}
		\centering
		\includegraphics[width=0.775\linewidth]{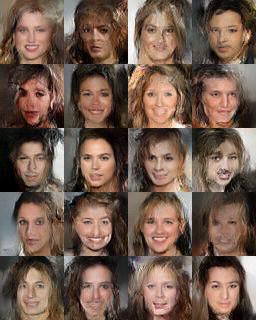}
		\caption*{Glow \\ Iteration 100,000}
	\end{subfigure}
	\begin{subfigure}[t]{0.31\textwidth}
		\centering
		\includegraphics[width=0.775\linewidth]{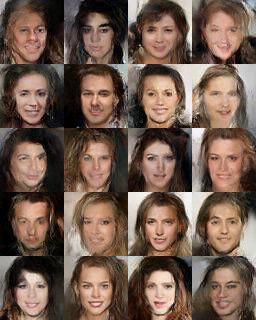}  
		\caption*{Glow \\ Iteration 600,000}
	\end{subfigure}
	\caption{Random samples $64 \times 64$ drawn from models trained on CelebA with temperature $0.7$.}
	\label{fig:celeba64-compare}
	\vspace{-0.5cm}
\end{figure*}

The test-set likelihoods are listed in Table~\ref{tab:nll} left. Our scores are worse than those reported by \citet{kingma2018glow,hoogeboom2019emerging} because we use smaller models and train each model on a single GPU. Based on the scores, 1$\times$1 convolutions perform the worst. Emerging convolutions and periodic convolutions score better than the 1$\times$1 convolutions, since they are more flexible and can model the dependencies along the spatial axes. Neural spline coupling works well on $32 \times 32$ images, but do slightly worse than 1$\times$1 convolution on $64 \times 64$ images. MaCow does not work well on ImageNet. This trend demonstrates the importance of permutation layers. They can model the interactions among dimensions and shuffle them, which coupling layers cannot do. Without a good permutation layer, a better coupling layer still cannot always improve the performance. The Woodbury transformation models perform the best, likely because they can model the interactions between the target dimension and all other dimensions, while the invertible convolutions only model the interactions between target dimension its neighbors. ME-Woodbury performs only slightly worse than the full version, showing that its restrictions provide a useful tradeoff between model quality and efficiency. 

We list model sizes in Table~\ref{tab:nll} (right). Despite modeling rich interactions, Woodbury transformations are not the largest. With $32 \times 32$ images, ME-Woodbury  and 1$\times$1 convolution are the same size. When the image size is $64 \times 64$, ME-Woodbury is the smallest. This is because we use the multi-scale architecture, i.e., Fig.~\ref{fig:flow-arc}, to combine layers. The squeeze layer doubles the input variable's channels at each level, so larger $L$ suggests larger $c$. The space complexities of invertible convolutions are $\mathcal{O}(c^2)$, while the space complexity of ME-Woodbury is linear to $c$. When $c$ is large, the weight matrices of invertible convolutions are larger than the weight matrices of ME-Woodbury.

\begin{wraptable}[10]{r}{0.4\textwidth}
	\vspace{-11pt}
	\caption{Evaluation of different $d$ (bpd).}
	\label{tab:d}
	\begin{tabular}{ l  l  l }
		\toprule
		~ & Woodbury & ME-Woodbury\\
		\midrule
		$d = 2$ & 3.54 & 3.53  \\
		$d = 4$ & 3.51 & 3.51 \\
		$d = 8$ & 3.48 & 3.48  \\
		$d = 16$ & 3.47 & 3.48 \\
		$d = 32$ & 3.47 & 3.48\\
		\bottomrule
	\end{tabular}
\end{wraptable}

\textbf{Latent Dimension Evaluation}~We test the impact of latent dimension $d$ on the performance of Woodbury-Glow. We train our models on CIFAR-10, and use bpd as metric. We vary $d$ within $\{2,4,8,16,32\}$. The results are in Table~\ref{tab:d}. When $d<8$, the model performance will be impacted. When $d>16$, increasing $d$ will not improve the bpd. This is probably because when $d$ is too small, the latent features cannot represent the input variables well, and when $d$ is too big, the models become hard to train. When $8 \le d \le 16$, the Woodbury transformations are powerful enough to model the interactions among dimensions. We also test two values of $d$, i.e., $16, 32$, of Woodbury-Glow on ImageNet $64 \times 64$. The bpds of both $d$ are $3.87$, which are consistent with our conclusion.

\begin{wrapfigure}[14]{r}{0.375\textwidth}
	\centering
	\vspace{-10pt}
	\includegraphics[width=1.0\linewidth]{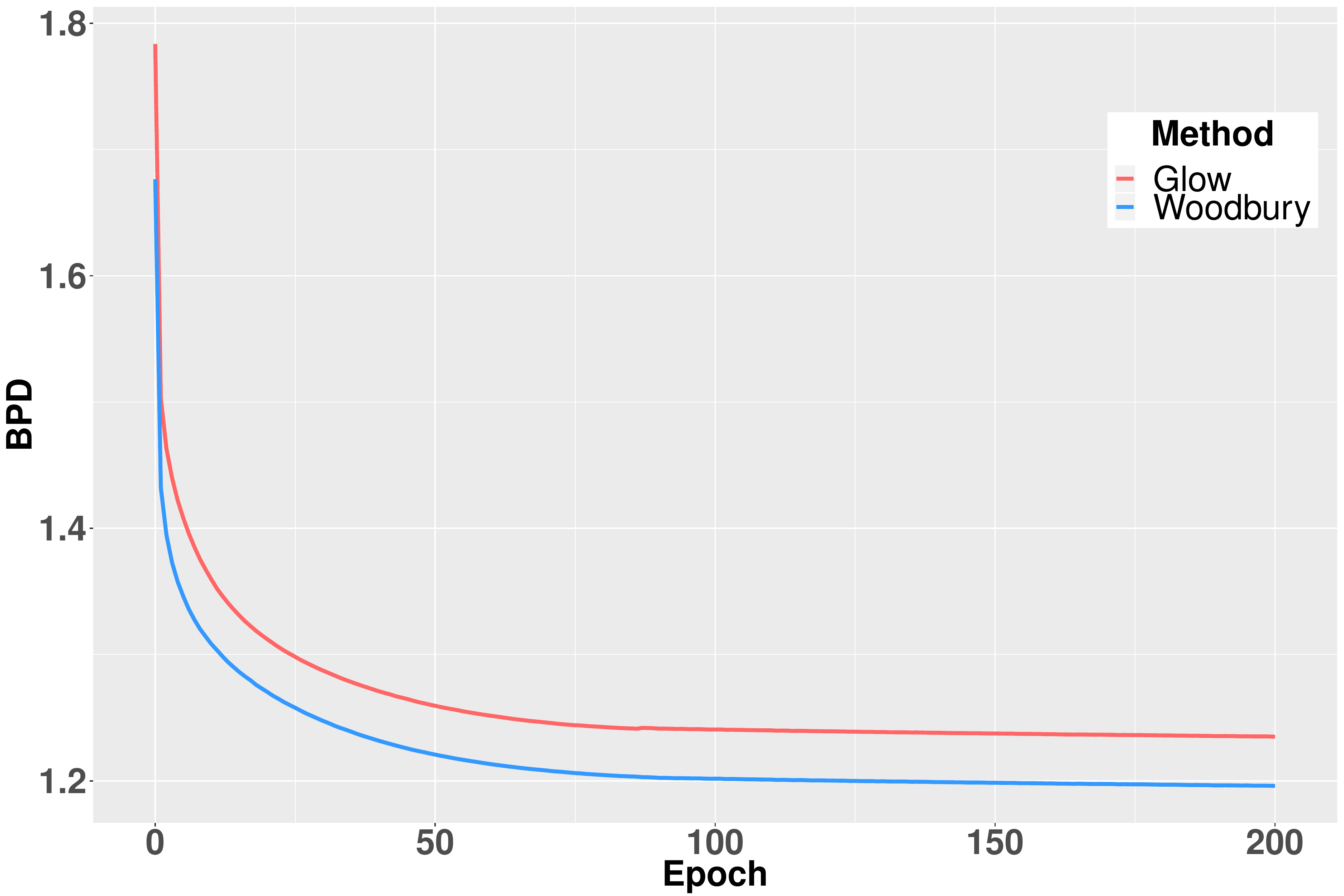}
	\caption{Learning curves on CelebA-HQ 64x64. The NLL of Woodbury Glow decreases faster than Glow.}
	\label{fig:learning_curves}
\end{wrapfigure}

\textbf{Sample Quality Comparisons}~
We train Glow and Woodbury-Glow on the CelebA-HQ dataset~\cite{karras2017progressive}. We use $5$-bit images and set the size of images to be $64 \times 64$, $128 \times 128$, and $256 \times 256$.  Due to our limited computing resources, we use relatively small models in our experiments. We follow \citet{kingma2018glow} and choose a temperature parameter to encourage higher quality samples. Detailed parameter settings are in the appendix. We compare samples from Glow and Woodbury-Glow during three phases of training, displayed in Fig.~\ref{fig:celeba64-compare}. The samples show a clear trend where Woodbury-Glow more quickly learns to generate reasonable face shapes. After 100,000 iterations, it can already generate reasonable samples, while Glow's samples are heavily distorted. Woodbury-Glow samples are consistently smoother and more realistic than samples from Glow in all phases of training. The samples demonstrate Woodbury transformations' advantages. The learning curves in Figure~\ref{fig:learning_curves} also show that the NLL of Woodbury Glow decreases faster, which is consistent to the sample comparisons.  In the appendix, we show analogous comparisons using higher resolution versions of CelebA data, which also exhibit the trend of Woodbury-Glow generating more realistic images than Glow at the same training iterations.

\section{Conclusion}
\label{sec:conclusion}

In this paper, we develop Woodbury transformations, which use the Woodbury matrix identity to compute the inverse transformations and Sylvester's determinant identity to compute Jacobian determinants. Our method has the same advantages as invertible $d \times d$ convolutions that can capture correlations among all dimensions. In contrast to the invertible $d \times d$ convolutions, our method is parallelizable and the computational complexity of our methods are linear to the input size, so that it is still efficient in computation when the input is high-dimensional. One potential limitation is that Woodbury transformations do not have parameter sharing scheme as in convolutional layers, so one potential future research is to develop partially Woodbury transformations that can share parameters. We test our models on multiple image datasets and they outperform state-of-the-art methods.

\section*{Broader Impact}

This paper presents fundamental research on increasing the expressiveness of deep probabilistic models. Its impact is therefore linked to the various applications of such models. 
By enriching the class of complex deep models for which we can train with exact likelihood, we may enable a wide variety of applications that can benefit from modeling of uncertainty. 
However, a potential danger of this research is that deep generative models have been recently applied to synthesize realistic images and text, which can be used for misinformation campaigns. 

\section*{Acknowledgments}
\label{sec:acknowledgments}
We thank NVIDIA's GPU Grant Program and Amazon's AWS Cloud Credits for Research program for their support. The work was completed while both authors were affiliated with the Virginia Tech Department of Computer Science. Bert Huang was partially supported by an Amazon Research Award and a grant from the U.S. Department of Transportation Safe-D Program for work on separate projects not directly related to this paper.


\newpage
\appendix

\section{More Background}
In this section, we introduce more detailed background knowledge.

\subsection{Normalizing Flows}

Let $\mathbf{x}$ be a high-dimensional continuous variable. We suppose that $\mathbf{x}$ is drawn from $p^*(\mathbf{x})$, which is the true data distribution. Given a collected dataset $\mathcal{D} = \{\mathbf{x}_1, \mathbf{x}_2,...,\mathbf{x}_D\}$, we are interested in approximating $p^*(\mathbf{x})$ with a model $p_{\theta}(\mathbf{x})$. We optimize $\theta$ by minimizing the negative log-likelihood
\begin{equation}
\mathcal{L}(\mathcal{D}) = \sum_{i=1}^{D} - \log p_{\theta}(\mathbf{x}_i).
\label{eq:likelihood}
\end{equation}

For some settings, variable $\tilde{\mathbf{x}}$ is discrete, e.g., image pixel values are often integers. In these cases, we dequantize $\tilde{\mathbf{x}}$ by adding continuous noise $\bm{\mu}$ to it, resulting in a continuous variable $\mathbf{x} = \tilde{\mathbf{x}} + \bm{\mu}$. As shown by \citet{ho2019flow}, the log-likelihood of $\tilde{\textbf{x}}$ is lower-bounded by the log-likelihood of $\mathbf{x}$. 

Normalizing flows enable computation of $p_{\theta}(\mathbf{x})$, even though it is usually intractable for many other model families. A normalizing flow~\cite{rezende2015variational} is composed of a series of invertible functions $\mathbf{f} = \mathbf{f}_1 \circ \mathbf{f}_2 \circ ... \circ \mathbf{f}_K$, which transform $\mathbf{x}$ to a latent code $\mathbf{z}$ drawn from a simple distribution. Therefore, with the \emph{change of variables} formula, we can rewrite $\log p_{\theta}(\mathbf{x})$ to be
\begin{equation}
\log p_{\theta}(\mathbf{x}) = \log p_{Z}(\mathbf{z}) + \sum_{i=1}^{K} \log \left|\det \left(\frac{\partial \mathbf{f}_i}{\partial \mathbf{r}_{i-1}}\right)\right|,
\label{eq:relikelihood}
\end{equation}
where $\mathbf{r}_i = \mathbf{f}_i(\mathbf{r}_{i-1})$, $\mathbf{r}_{0} = \mathbf{x}$, and $\mathbf{r}_{K}=\mathbf{z}$. 

\subsection{Invertible $d \times d$ Convolutions}

Emerging convolutions~\citep{hoogeboom2019emerging} combine two autoregressive convolutions~\citep{germain2015made,kingma2016improved}. Formally, 
\begin{eqnarray*}
	\mathbf{M}'_1 = \mathbf{M}_1 \odot \mathbf{A}_1, ~~~~~~~~
	\mathbf{M}'_2 = \mathbf{M}_2 \odot \mathbf{A}_2, ~~~~~~~~
	\mathbf{y} = \mathbf{M}'_2 \star (\mathbf{M}'_1 \star \mathbf{x}),
\end{eqnarray*}
where $\mathbf{M}_1, \mathbf{M}_2$ are convolutional kernels whose size is $c \times c \times d \times d$, and $\mathbf{A}_1, \mathbf{A}_2$ are binary masks. The symbol $\star$ represents the convolution operator.\footnote{In practice, a convolutional layer is usually implemented as an aggregation of cross-correlations. We follow \citet{hoogeboom2019emerging} and omit this detail.} An emerging convolutional layer has the same receptive fields as standard convolutional layers, which can capture correlations between a target pixel and its neighbor pixels. However, like other autoregressive convolutions, computing the inverse of an emerging convolution requires sequentially traversing each dimension of input, so its computation is not parallelizable and is a computational bottleneck when the input is high-dimensional.

Periodic convolutions~\cite{hoogeboom2019emerging,Finzi2019Invertible} use discrete Fourier transformations to transform both the input and the kernel to Fourier domain. A periodic convolution is computed as
\begin{equation*}
\mathbf{y}_{u,:,:} =  \sum_{v} \mathcal{F}^{-1}(\mathcal{F}(\mathbf{M}^{(p)}_{u,v,:,:})\odot \mathcal{F}(\mathbf{x}_{v,:,:})),
\end{equation*}
where $\mathcal{F}$ is a discrete Fourier transformation, and $\mathbf{M}^{(p)}$ is the convolution kernel whose size is $c \times c \times d \times d$.  The computational complexity of periodic convolutions is $\mathcal{O}(c^2hw\log(hw) +c^3hw)$. In our experiments, we found that the Fourier transformation requires a large amount of memory. These two problems impact the efficiency of both training and sampling when the input is high-dimensional.

\section{Memory-Efficient Woodbury Transformations}
\label{sec:me-woodbury-appendix}

Memory-Efficient Woodbury transformations can effectively reduce the space complexity. The main idea is to perform spatial transformations along the height and width axes separately, i.e., a height transformation and a width transformation. The transformations are:
\begin{eqnarray}
\mathbf{x}_c &=& (\mathbf{I}^{(c)} + \mathbf{U}^{(c)}\mathbf{V}^{(c)}) \mathbf{x}, \nonumber\\
\mathbf{x}_w &=& \text{reshape}(\mathbf{x}_c, (ch, w)), \nonumber\\
\mathbf{x}_w &=& \mathbf{x}_c (\mathbf{I}^{(w)} + \mathbf{U}^{(w)}\mathbf{V}^{(w)}), \nonumber\\
\mathbf{x}_h &=& \text{reshape}(\mathbf{x}_w, (cw, h)), \nonumber \\
\mathbf{y} &=& \mathbf{x}_h(\mathbf{I}^{(h)} + \mathbf{U}^{(h)}\mathbf{V}^{(h)}), \nonumber\\
\mathbf{y} &=& \text{reshape}(\mathbf{y}, (c, hw)),
\label{eq:me-w}
\end{eqnarray}
where $\text{reshape}(\mathbf{x}, (n,m))$ reshapes $\mathbf{x}$ to be an $n \times m$ matrix. Matrices $\mathbf{I}^{(w)}$ and $\mathbf{I}^{(h)}$ are $w$- and $h$-dimensional identity matrices, respectively. Matrices $\mathbf{U}^{(w)}, \mathbf{V}^{(w)}, \mathbf{U}^{(h)}$, and $\mathbf{V}^{(h)}$ are $w \times d_w$, $d_w \times w$, $w \times d_w$, and $d_w \times w$ matrices, respectively, where $d_w$ and $d_h$ are constant latent dimensions. 

Using the Woodbury matrix identity and the Sylvester's determinant identity, we can compute the inverse and Jacobian determinant:
\begin{eqnarray}
\mathbf{y} &=& \text{reshape}(\mathbf{y}, (cw, h)), \nonumber\\
\mathbf{x}_h &=& \mathbf{y}(\mathbf{I}^{(h)} - \mathbf{U}^{(h)}(\mathbf{I}^{(d_h)} + \mathbf{V}^{(h)}\mathbf{U}^{(h)})^{-1}\mathbf{V}^{(h)}), \nonumber\\
\mathbf{x}_w &=& \text{reshape}(\mathbf{x}_h, (ch, w)), \nonumber\\
\mathbf{x}_w &=& \mathbf{x}_w(\mathbf{I}^{(w)} - \mathbf{U}^{(w)}(\mathbf{I}^{(d_w)} + \mathbf{V}^{(w)}\mathbf{U}^{(w)})^{-1}\mathbf{V}^{(w)}), \nonumber \\
\mathbf{x}_c &=& \text{reshape}(\mathbf{x}_w, (c, hw)), \nonumber\\
\mathbf{x} &=& (\mathbf{I}^{(c)} - \mathbf{U}^{(c)}(\mathbf{I}^{(d_c)} + \mathbf{V}^{(c)}\mathbf{U}^{(c)})^{-1}\mathbf{V}^{(c)})\mathbf{x}_c,
\end{eqnarray}
\begin{eqnarray}
\log \left| \det(\frac{\partial \mathbf{y}}{\partial \mathbf{x}}) \right|  &= hw \log\left|\det(\mathbf{I}^{(d_c)}+\mathbf{V}^{(c)}\mathbf{U}^{(c)})\right| + ch\log\left|\det(\mathbf{I}^{(d_w)}+\mathbf{V}^{(w)}\mathbf{U}^{(w)})\right| \nonumber\\
&+ cw\log\left|\det\left(\mathbf{I}^{(d_h)}+\mathbf{V}^{(h)}\mathbf{U}^{(h)}\right)\right|,
\end{eqnarray}
where $\mathbf{I}^{(d_w)}$ and $\mathbf{I}^{(d_h)}$ are $d_w$- and $d_h$-dimensional identity matrices, respectively. The Jacobian of the $\text{reshape}()$ is an identity matrix, so its log-determinant is $0$. 

We call Equation~\ref{eq:me-w} the memory-efficient Woodbury transformation because it reduces space complexity from $\mathcal{O}(c+hw)$ to $\mathcal{O}(c+h+w)$. This method is effective when $h$ and $w$ are large. To analyze its complexity, we let all latent dimensions be less than $d$ as before. The complexity of forward transformation is $\mathcal{O}(dchw)$; the complexity of computing the determinant is $\mathcal{O}(d(c+h+w)+d^3)$; and the complexity of computing the inverse is $\mathcal{O}(dchw + d^2(c+ch+cw)+d^3)$. The same as Woodbury transformations, when the input is high dimensional, we can omit $d$. Therefore, the computational complexities of the memory-efficient Woodbury transformation are also linear with the input size. 

We list the complexities of different methods in Table~\ref{tab:complexity}. We can see that the computational complexities of Woodbury transformations are comparable to other methods, and maybe smaller when the input is high-dimensional, i.e., the $c,h,w$ are big.

\begin{table}[!htp]
	\begin{center}
		\caption{Comparisons of computational complexities. }
		\label{tab:complexity}
		\begin{tabular}{lll}
			\toprule
			Method & Forward &Backward\\
			\midrule
			1x1 convolution & $\mathcal{O}(c^2hw+c^3)$ &  $\mathcal{O}(c^2hw)$  \\
			Periodic conolution & $\mathcal{O}(chw\log(hw)+c^3hw)$ &  $\mathcal{O}(chw\log(hw)+c^2hw)$  \\
			Emerging convolution & $\mathcal{O}(c^2hw)$ & $\mathcal{O}(c^2hw)$ \\
			ME-Woodbury transformation & $\mathcal{O}(dchw)$  & $\mathcal{O}(dchw)$   \\
			Woodbury transformation & $\mathcal{O}(dchw)$ & $\mathcal{O}(dchw)$ \\
			\bottomrule
		\end{tabular}
	\end{center}
\end{table}

\section{Parameter Settings}
\label{sec:parametersetting-appendix}
In this section, we present additional details about our experiments to aid reproducibility.

\subsection{Experiments of Quantitative Evaluation}

In the experiments of qualitative evalution, we compare Woodbury transformations with $3$ permutation layer baselines, i.e., 1x1 convolution, emerging convolution, and periodic coupling, and $2$ coupling layer baselines, i.e., neural spline coupling, and MaCow. For all generalized permutation methods, we use affine coupling, which is composed of $3$ convolutional layers, and the $2$ latent layers have $512$ channels. For the neural spline coupling, we set the number of spline bins to $4$. The spline parameters are generated by a neural network, which is also composed of convolutional layers. For $32 \times 32$ images, we set the number of channels to $256$, and for $64 \times 64$ images, we set it to $224$. \citet{macow2019} used steps containing a MaCow unit, i.e., $4$ autoregressive convolution coupling layers, and a full Glow step. For fair comparison, we directly use the MaCow unit to replace the affine coupling. For $32 \times 32$ images, we set the convolution channel to $384$, and for $64 \times 64$ images, we set it to $296$.

We run each method to fixed number of iterations and test it every $10,000$ iterations. The bpds reported in our main paper are the best bpds obtained by each method. The bpds are single-run results. This is because each run of the experiment requires 3 to 5 days, and running each model multiple times is a major cost. We found in our experiments that for the same model and parameter settings, the bpds' standard deviation of multiple runs are very small, i.e., around $0.003$, so single run results are sufficient for comparing bpd. 

\subsection{Hyper-parameter Settings}

We use Adam~\cite{kingma2014adam} to tune the learning rates, with $\alpha=0.001$, $\beta_1=0.9$, and $\beta_2 = 0.999$. We use uniform dequantization. The sizes of models we use, and mini-batch sizes for training in our experiments are listed in Table~\ref{tab:hyper}. 

\begin{table}[!htp]
	\begin{center}
		\caption{Model sizes and mini-batch sizes. }
		\label{tab:hyper}
		\begin{tabular}{lllll}
			\toprule
			Dataset & Mini-batch size & Levels(L) & Steps(K) & Coupling channels \\
			\midrule
			CIFAR-10 32x32 & 64 &  3 & 8 & 512 \\
			ImageNet 32x32 & 64 & 3 & 8 & 512 \\
			ImageNet 64x64 & 32 & 4 & 16  & 512\\
			LSUN Church 96x96 & 16 & 5 & 16 & 256 \\
			CelebA-HQ 64x64 & 8 & 4 & 16 & 512 \\
			CelebA-HQ 128x128 & 4 & 5 & 24 & 256 \\
			CelebA-HQ 256x256 & 4 & 6 & 16 & 256 \\
			\bottomrule
		\end{tabular}
	\end{center}
\end{table}

\subsection{Latent Dimension Settings}

%

In all our experiments, we set the latent dimensions of Woodbury transformations, and ME-Woodbury transformations as in Table~\ref{tab:w-latent}. 

\begin{table}[!htp]
	\begin{center}
		\caption{Latent dimensions of Woodbury transformations and ME-Woodbury transformations. The numbers in the brackets represent the latent dimension used in that level. For example, the  $d_c: \{8, 8, 16\}$, represents that the settings of $d_c$ at the three levels are $8$, $8$, and $16$. }
		\label{tab:w-latent}
		\begin{tabular}{lll}
			\toprule
			Dataset & Woodbury & ME-Woodbury \\
			\midrule
			CIFAR-10 32x32 & $d_c: \{8, 8, 16\}$ & $d_c: \{8, 8, 16\}$ \\
			~ & $d_s: \{16, 16, 8\}$ & $d_h: \{16, 16, 8\}$  \\
			~ & ~ & $ d_w: \{16, 16, 8\}$ \\
			\addlinespace[0.5em]
			ImageNet 32x32 & $d_c: \{8, 8, 16\}$ & $d_c: \{8, 8, 16\}$  \\
			~ & $d_s: \{16, 16, 8\}$ & $d_h: \{16, 16, 8\}$ \\
			~ & ~& $d_w: \{16, 16, 8\}$  \\
			\addlinespace[0.5em]
			ImageNet 64x64 & $d_c: \{8, 8, 16, 16\}$ & $d_c: \{8, 8, 16, 16\}$  \\
			~ & $d_s: \{16, 16, 8, 8\}$ &  $d_h: \{16, 16, 8, 8\}$ \\
			~ & ~ & $d_w: \{16, 16, 8, 8\}$ \\
			\addlinespace[0.5em]
			LSUN Church 96x96 & $d_c: \{8, 8, 16, 16, 16\}$ & --- \\
			~ & $d_s: \{16, 16, 16, 8, 8\}$ \\
			\addlinespace[0.5em]
			CelebA-HQ 64x64& $d_c: \{8, 8, 16, 16\}$ & ---\\
			~ & $d_s: \{16, 16, 8, 8\}$ \\
			\addlinespace[0.5em]
			CelebA-HQ 128x128 & $d_c: \{8, 8, 16, 16, 16\}$ & ---\\
			~ & $d_s: \{16, 16,16, 8, 8\}$ \\
			\addlinespace[0.5em]
			CelebA-HQ 256x256 & $d_c: \{8, 8, 16, 16,16,16\}$ & --- \\
			~ & $d_s: \{16, 16, 16, 16,  8, 8\}$ \\
			\bottomrule
		\end{tabular}
	\end{center}
\end{table}
\clearpage

\section{Sample Quality Comparisons}

\begin{figure*}[tbp]
	\centering
	\begin{subfigure}[t]{0.31\textwidth}
		\centering
		\includegraphics[width=1.0\linewidth]{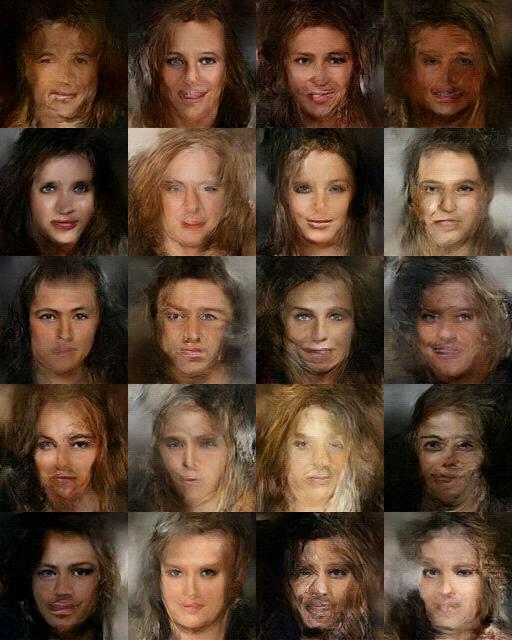}
		\caption*{Woodbury-Glow \\ Iteration 100,000}
	\end{subfigure}
	\begin{subfigure}[t]{0.31\textwidth}
		\centering
		\includegraphics[width=1.0\linewidth]{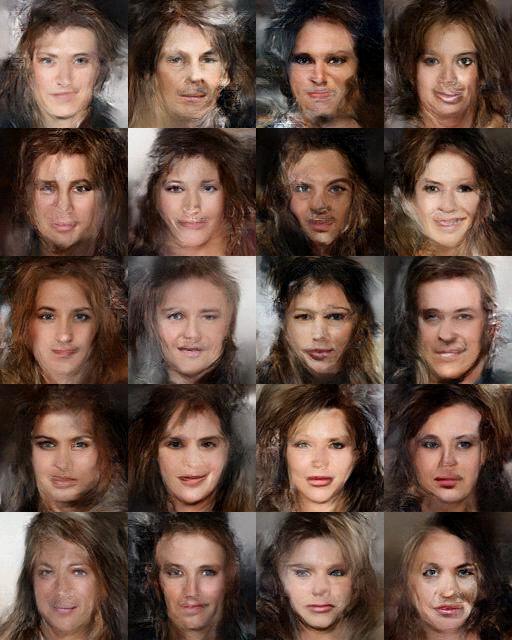}  
		\caption*{Woodbury-Glow \\ Iteration 200,000}
	\end{subfigure}
	\begin{subfigure}[t]{0.31\textwidth}
		\centering
		\includegraphics[width=1.0\linewidth]{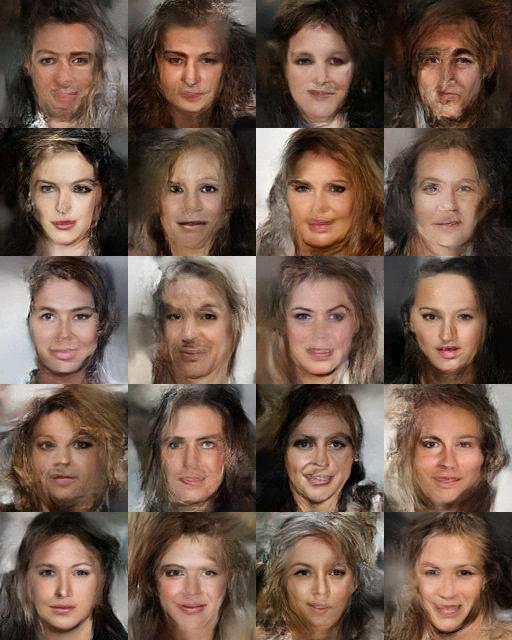}
		\caption*{Woodbury-Glow \\ Iteration 300,000}
	\end{subfigure}
	\begin{subfigure}[t]{0.31\textwidth}
		\centering
		\includegraphics[width=1.0\linewidth]{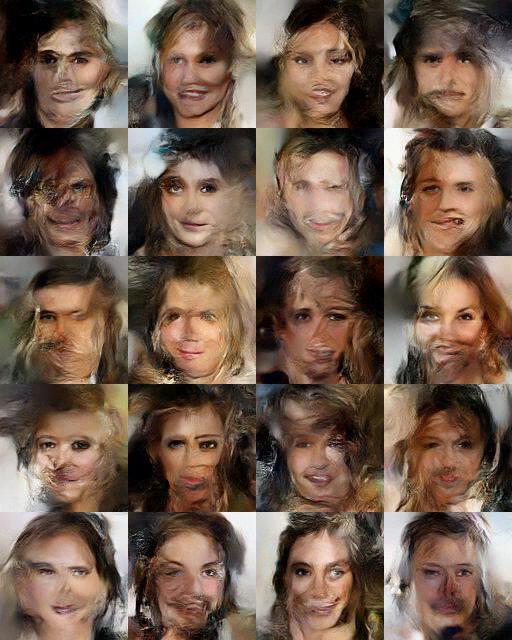}  
		\caption*{Glow \\ Iteration 100,000}
	\end{subfigure}
	\begin{subfigure}[t]{0.31\textwidth}
		\centering
		\includegraphics[width=1.0\linewidth]{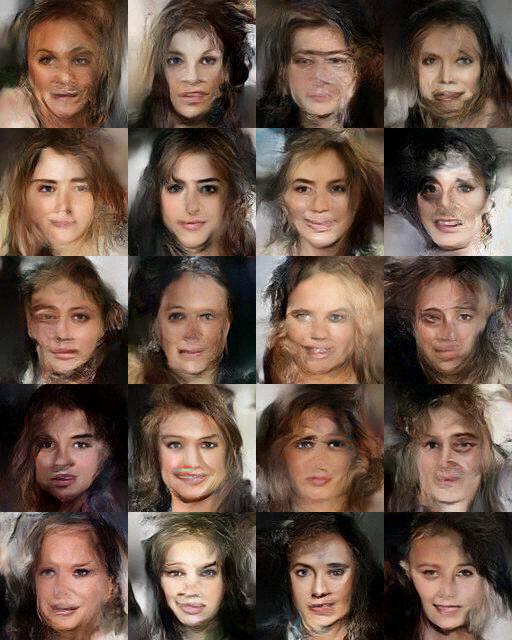}
		\caption*{Glow \\ Iteration 200,000}
	\end{subfigure}
	\begin{subfigure}[t]{0.31\textwidth}
		\centering
		\includegraphics[width=1.0\linewidth]{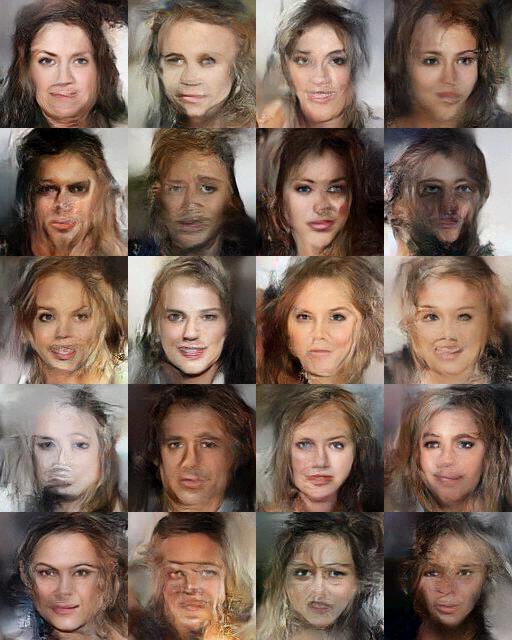}  
		\caption*{Glow \\ Iteration 300,000}
	\end{subfigure}
	\caption{Random samples of $128 \times 128$ images drawn with temperature $0.7$ from a model trained on CelebA data.}
	\label{fig:celeba128-compare}
\end{figure*}

\begin{figure*}[tbp]
	\centering
	\begin{subfigure}[t]{0.31\textwidth}
		\centering
		\includegraphics[width=1.0\linewidth]{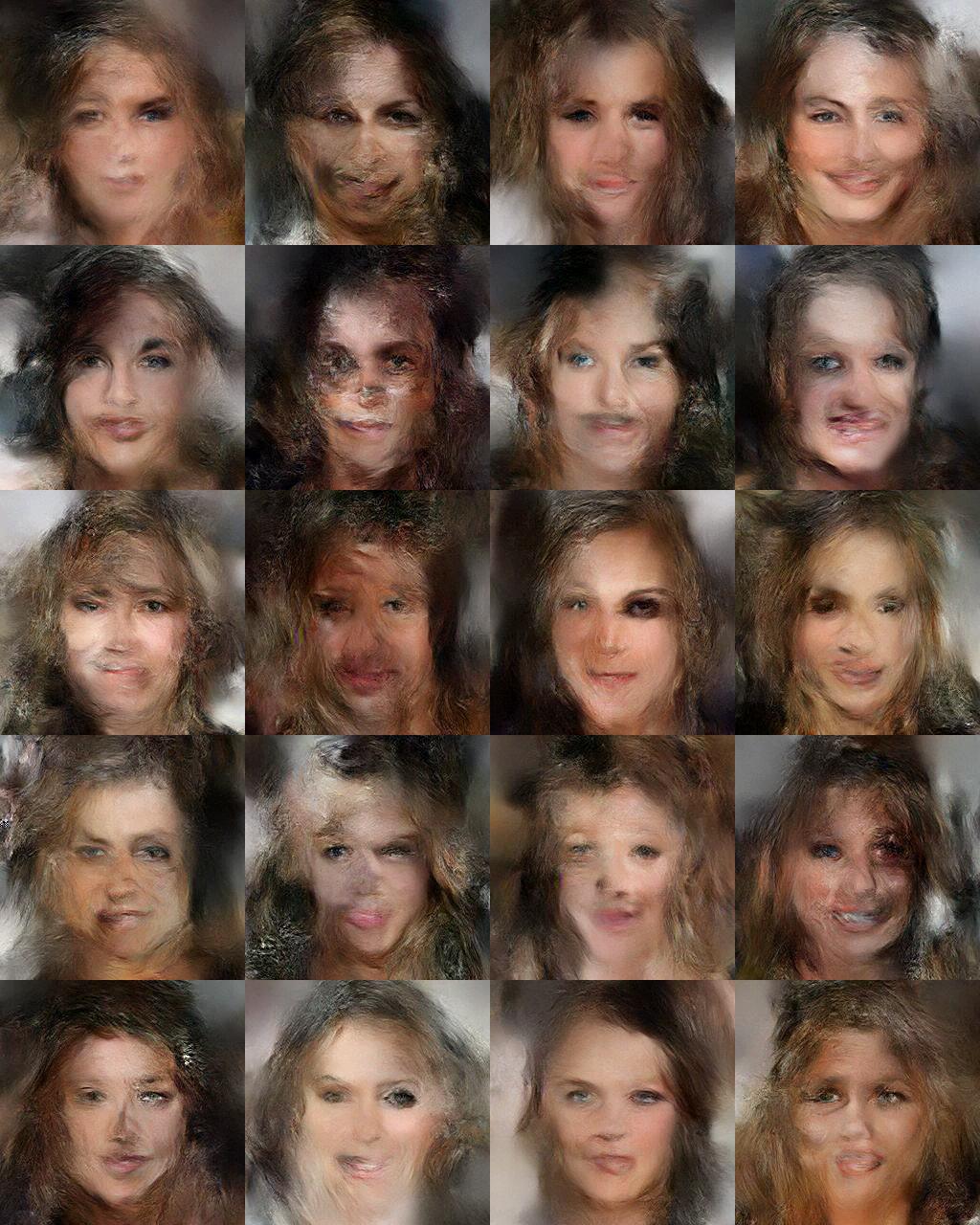}
		\caption*{Woodbury-Glow \\ Iteration 150,000}
	\end{subfigure}
	\begin{subfigure}[t]{0.31\textwidth}
		\centering
		\includegraphics[width=1.0\linewidth]{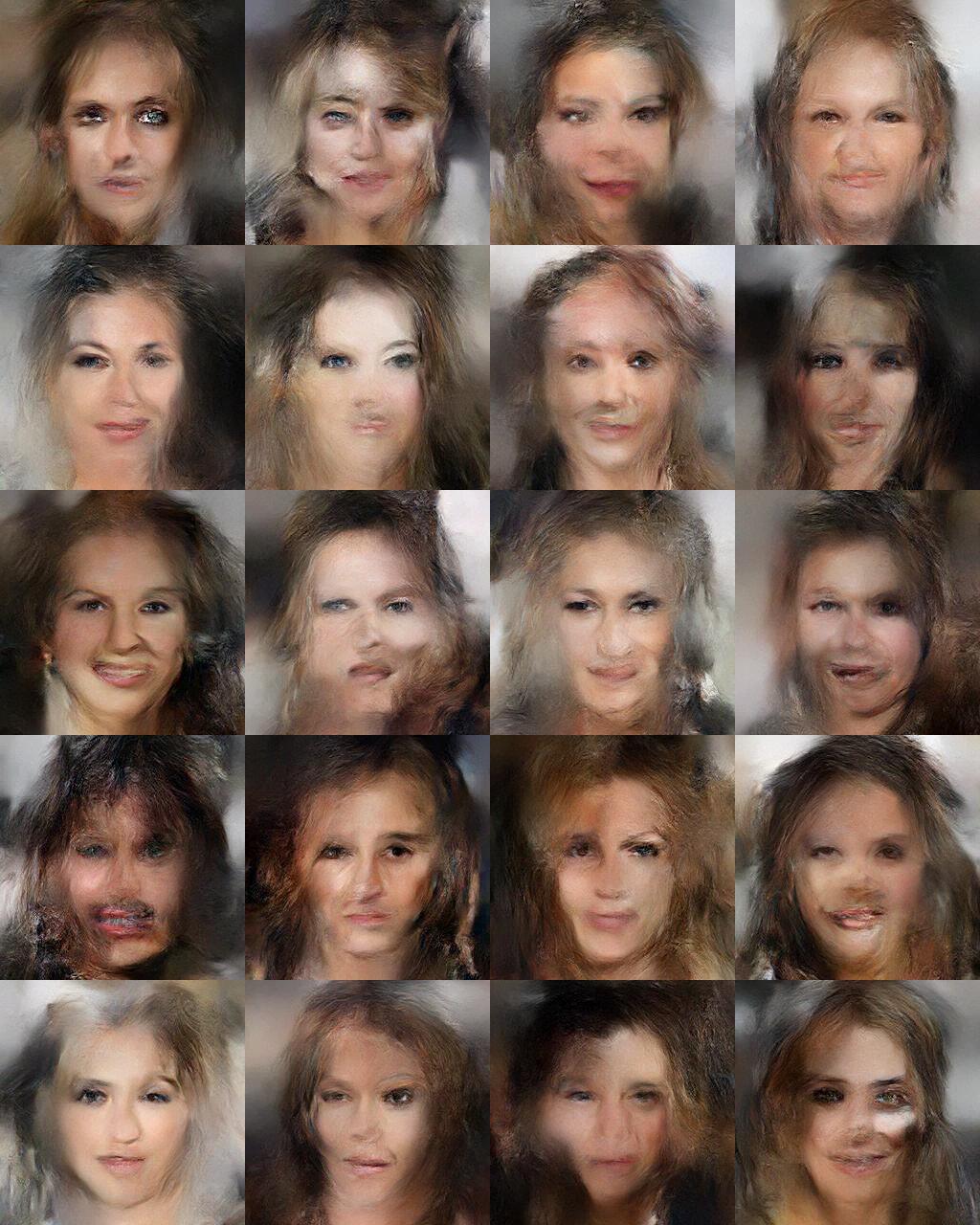}  
		\caption*{Woodbury-Glow \\ Iteration 220,000}
	\end{subfigure}
	\begin{subfigure}[t]{0.31\textwidth}
		\centering
		\includegraphics[width=1.0\linewidth]{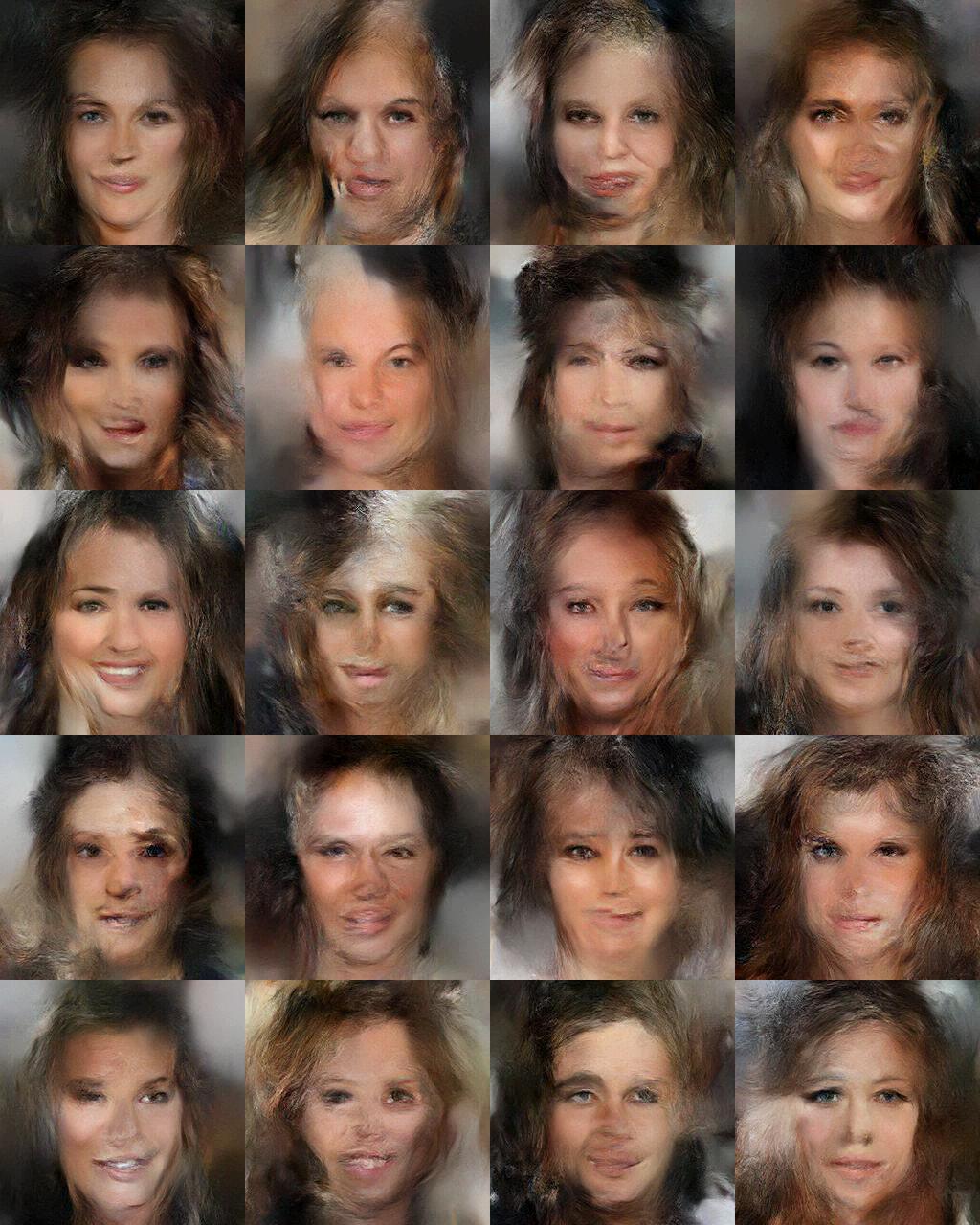}
		\caption*{Woodbury-Glow \\ Iteration 300,000}
	\end{subfigure}
	\begin{subfigure}[t]{0.31\textwidth}
		\centering
		\includegraphics[width=1.0\linewidth]{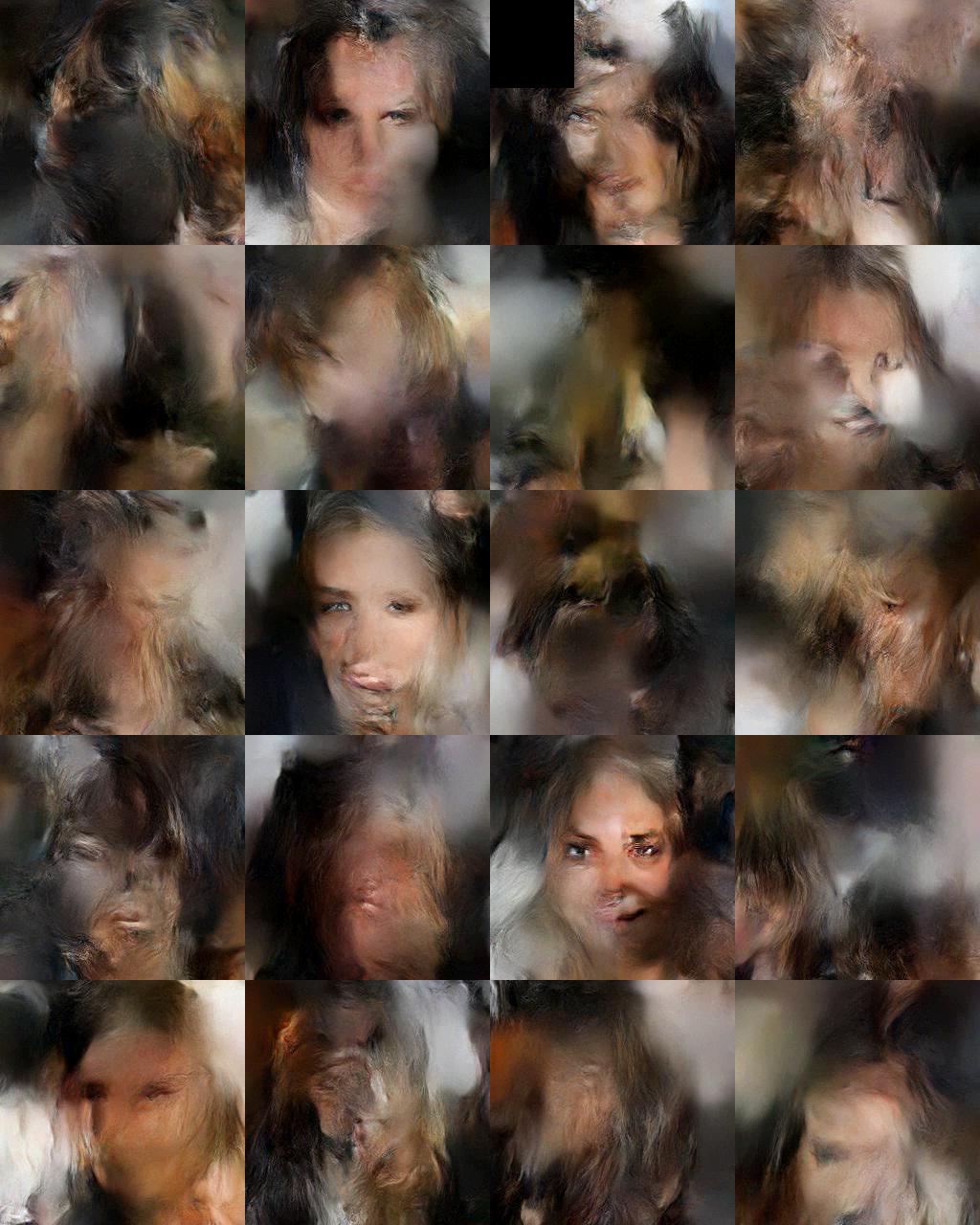}  
		\caption*{Glow \\ Iteration 150,000}
	\end{subfigure}
	\begin{subfigure}[t]{0.31\textwidth}
		\centering
		\includegraphics[width=1.0\linewidth]{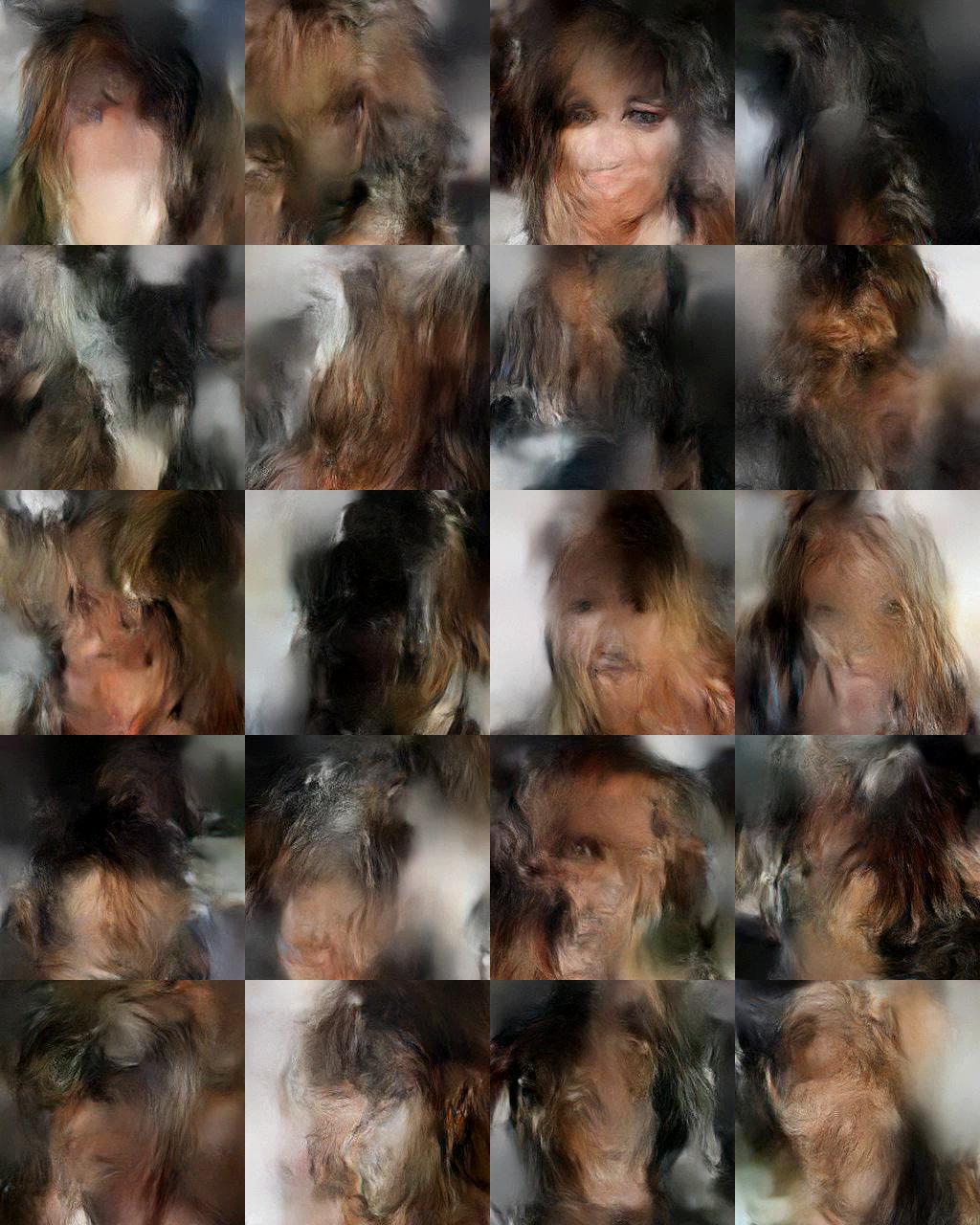}
		\caption*{Glow \\ Iteration 220,000}
	\end{subfigure}
	\begin{subfigure}[t]{0.31\textwidth}
		\centering
		\includegraphics[width=1.0\linewidth]{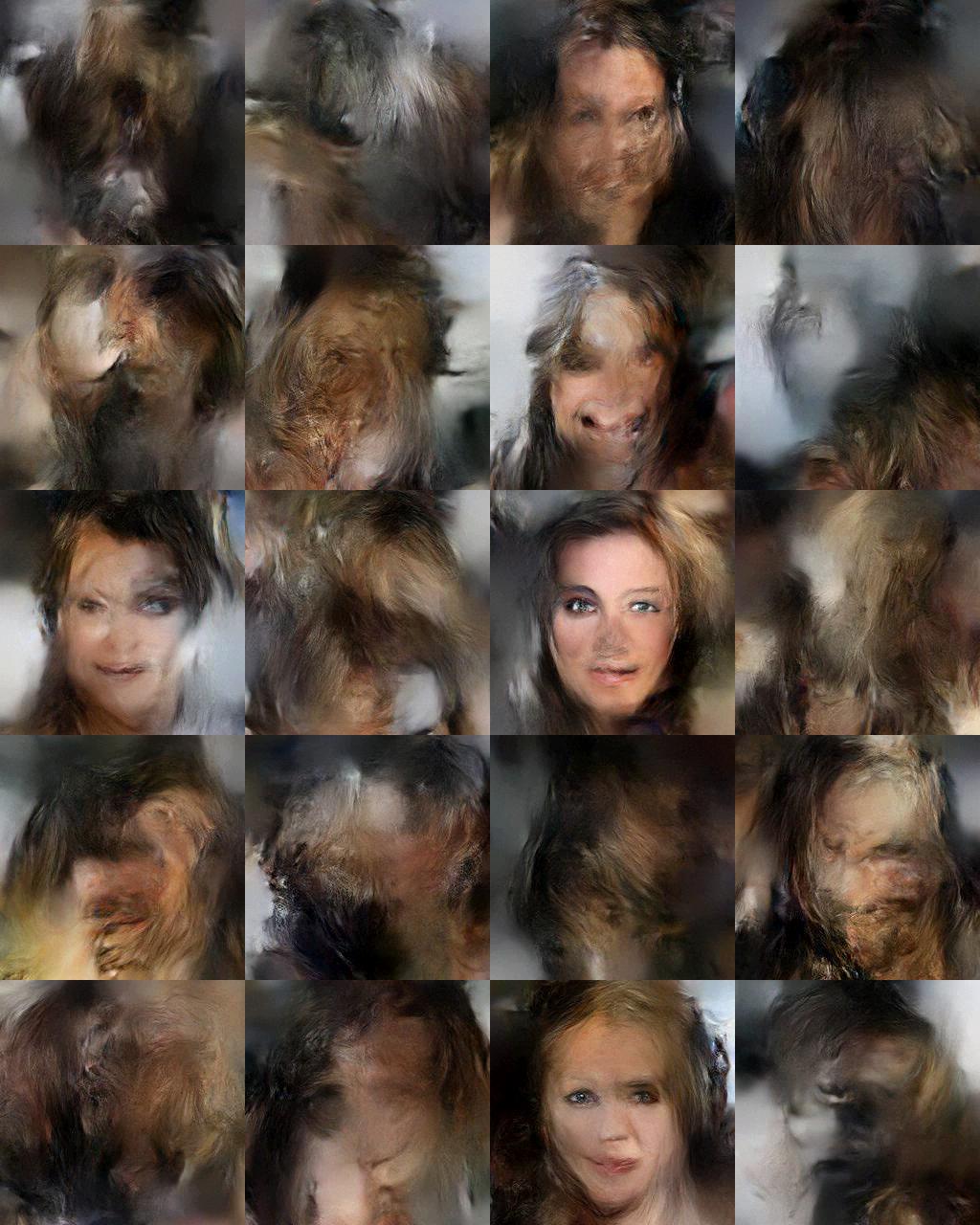}  
		\caption*{Glow \\ Iteration 300,000}
	\end{subfigure}
	\caption{Random samples of $256 \times 256$ images drawn with temperature $0.7$ from a model trained on CelebA data.}
	\label{fig:celeba256-compare}
\end{figure*}

We compare the samples generated by Woodbury-Glow and Glow models trained on the CelebA-HQ dataset. We follow \citet{kingma2018glow} and randomly hold out 3,000 images as a test set. We use $5$-bits images. We use $64 \times 64$, $128 \times 128$, $256 \times 256$ images. Due to the our limited computing resources, we use relatively small models. The model sizes and other settings are listed in Table~\ref{tab:hyper} and Table~\ref{tab:w-latent}. We generate samples from the models during different phases of training and display them in Figure~\ref{fig:celeba128-compare}, and Figure~\ref{fig:celeba256-compare} (The results of $64 \times 64$ images are shown in the main paper). 
For the $128 \times 128$ images, both Glow and Woodbury-Glow generate distorted images at iteration 100,000, but Woodbury-Glow seems to improve in later stages, stabilizing the shapes of faces and structure of facial features. Glow, continues generating faces with distorted overall shapes as training continues. 
For the $256 \times 256$ images, neither model ever trains sufficiently to generate highly realistic faces, but Woodbury-Glow makes significantly more progress in these 300,000 iterations than Glow. Glow's samples at 300,000 are still mostly random swirls with an occasional recognizable face, while almost all of Woodbury-Glow's samples look like faces, though distorted. Due to limits on our computational resources, we stopped the higher resolution experiments at 300,000 iterations (rather than running to 600,000 iterations as we did for the $64 \times 64$ experiments in the main paper). With a larger model and longer training time, it seems Woodbury-Glow would reach higher sample quality much faster than Glow.

The likelihoods of test set under the trained model are listed in Table~3. For the $64 \times 64$ and $128 \times 128$ images, Woodbury-Glow scores higher likelihood than Glow. For the $256 \times 256$ images, their likelihoods are almost identical, and are better than the score reported in \cite{kingma2018glow}. This may be due to three possible reasons: (1) We use affine coupling rather than additive coupling, which is a non-volume preserving layer and may improve the likelihoods; (2) Since the test set is randomly collected, it is different from the one used in \cite{kingma2018glow}; And (3) The model used in \cite{kingma2018glow} is very large, so it may be somewhat over-fitting. Surprisingly, the clear difference in sample quality is not reflected by the likelihoods. This discrepancy may be because we use $5$-bit images, and the images are all faces, so the dataset is less complicated than other datasets such as ImageNet. Moreover, even though Glow cannot generate reasonable $256 \times 256$ samples, the colors of these samples already match the colors of real images well, so these strange samples may non-intuitively be equivalently likely as the face-like samples from Woodbury-Glow.


\FloatBarrier

\begin{table}[ht]
	\begin{center}
		\caption{Bit per-dimension results on CelebA-HQ}
		\begin{tabular}{ l  l  l  }
			\toprule
			Size of images & Glow & Woodbury-Glow \\
			\midrule
			$64 \times 64$ & 1.27 & \textbf{1.23} \\
			$128 \times 128$ & 1.09 & \textbf{1.04} \\
			$256 \times 256$ & \textbf{0.93} & \textbf{0.93} \\
			\bottomrule
		\end{tabular}
	\end{center}
	\label{tab:celeba-nll}
\end{table}

\FloatBarrier

\section{Additional Samples}

\begin{figure}[!htp]
	\centering
	\begin{subfigure}[t]{0.45\textwidth}
		\centering
		\includegraphics[width=1.0\linewidth]{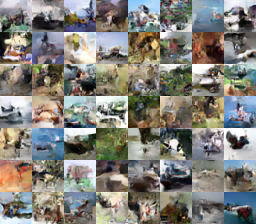}
		\caption*{}
	\end{subfigure}
	\begin{subfigure}[t]{0.45\textwidth}
		\centering
		\includegraphics[width=1.0\linewidth]{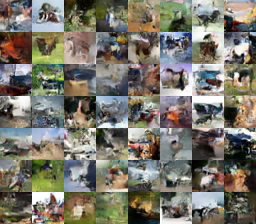}  
		\caption*{}
	\end{subfigure}
	\caption{CIFAR-10 $32 \times 32$ Woodbury-Glow samples.}
	\label{fig:cifar-sample}
\end{figure}

\begin{figure}[tbp]
	\centering
	\begin{subfigure}[t]{0.45\textwidth}
		\centering
		\includegraphics[width=1.0\linewidth]{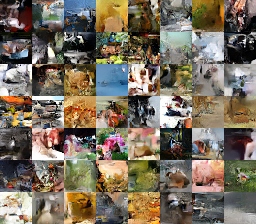}
		\caption*{}
	\end{subfigure}
	\begin{subfigure}[t]{0.45\textwidth}
		\centering
		\includegraphics[width=1.0\linewidth]{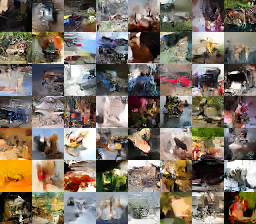}
		\caption*{}
	\end{subfigure}
	\caption{ImageNet $32 \times 32$ Woodbury-Glow samples.}
	\label{fig:imagenet32-sample}
\end{figure}

\begin{figure*}[tbp]
	\centering
	\begin{subfigure}[t]{0.45\textwidth}
		\centering
		\includegraphics[width=1.0\linewidth]{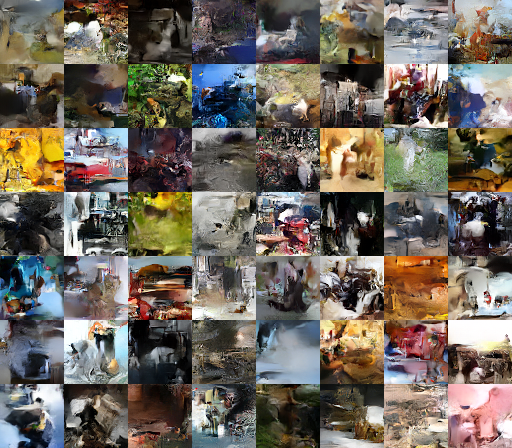}
		\caption*{}
	\end{subfigure}
	\begin{subfigure}[t]{0.45\textwidth}
		\centering
		\includegraphics[width=1.0\linewidth]{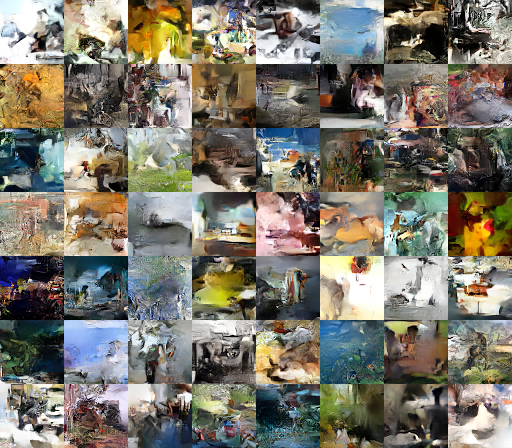}  
		\caption*{}
	\end{subfigure}
	\caption{ImageNet $64 \times 64$ Woodbury-Glow samples.}
	\label{fig:imagenet64-sample}
\end{figure*}

\begin{figure*}[tbp]
	\centering
	\begin{subfigure}[t]{0.45\textwidth}
		\centering
		\includegraphics[width=1.0\linewidth]{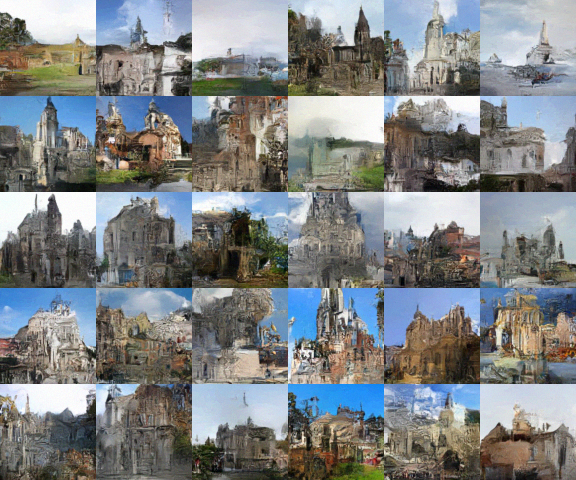}
		\caption*{}
	\end{subfigure}
	\begin{subfigure}[t]{0.45\textwidth}
		\centering
		\includegraphics[width=1.0\linewidth]{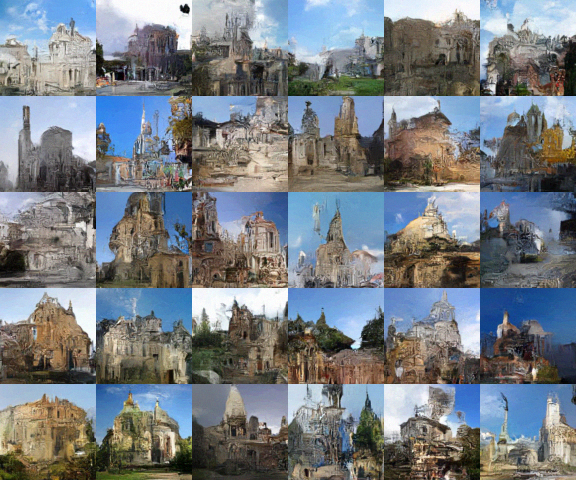}  
		\caption*{}
	\end{subfigure}
	\caption{LSUN church $96 \times 96$ Woodbury-Glow samples (temperature $0.875$).}
	\label{fig:lsun-sample}
\end{figure*}

\begin{figure*}[tbp]
	\centering
	\begin{subfigure}[t]{0.45\textwidth}
		\centering
		\includegraphics[width=1.0\linewidth]{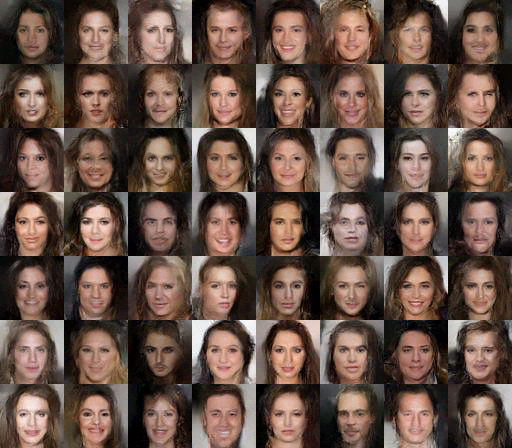}
		\caption*{}
	\end{subfigure}
	\begin{subfigure}[t]{0.45\textwidth}
		\centering
		\includegraphics[width=1.0\linewidth]{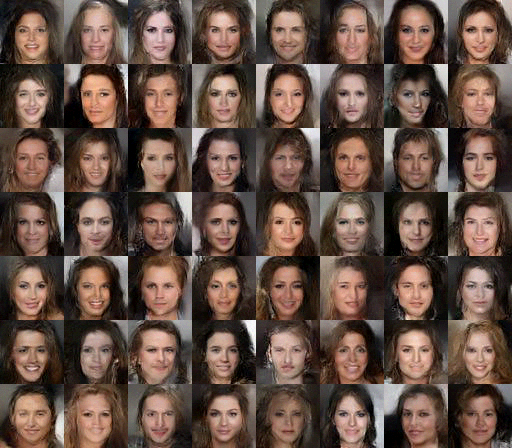}  
		\caption*{}
	\end{subfigure}
	\caption{CelebA-HQ $64 \times 64$ Woodbury-Glow samples (temperature $0.7$).}
	\label{fig:celeba64-sample}
\end{figure*}

\begin{figure*}[tbp]
	\centering
	\begin{subfigure}[t]{0.45\textwidth}
		\centering
		\includegraphics[width=1.0\linewidth]{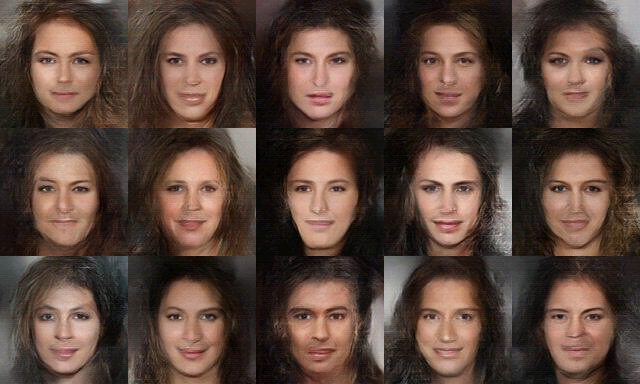}
		\caption*{}
	\end{subfigure}
	\begin{subfigure}[t]{0.45\textwidth}
		\centering
		\includegraphics[width=1.0\linewidth]{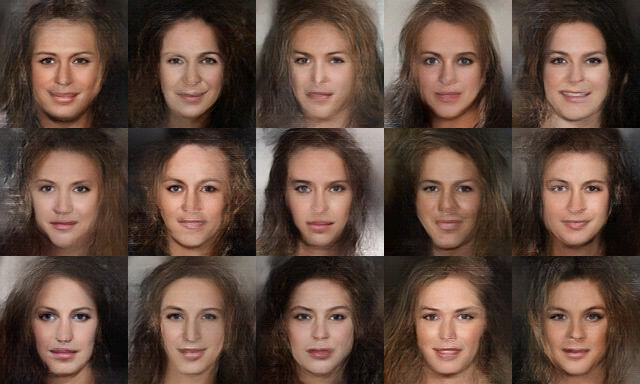}  
		\caption*{}
	\end{subfigure}
	\caption{CelebA-HQ $128 \times 128$ Woodbury-Glow samples (temperature $0.5$).}
	\label{fig:celeba128-sample}
\end{figure*}

\begin{figure*}[tbp]
	\centering
	\begin{subfigure}[t]{0.45\textwidth}
		\centering
		\includegraphics[width=1.0\linewidth]{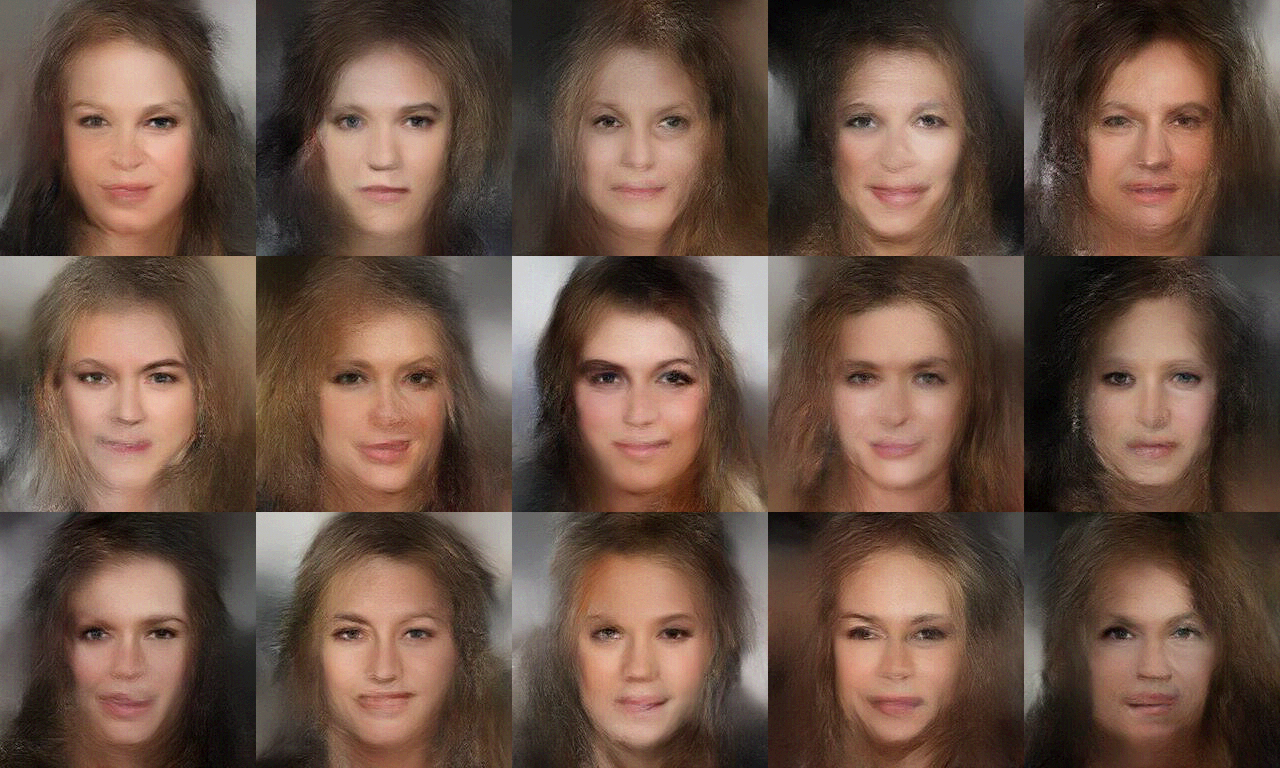}
		\caption*{}
	\end{subfigure}
	\begin{subfigure}[t]{0.45\textwidth}
		\centering
		\includegraphics[width=1.0\linewidth]{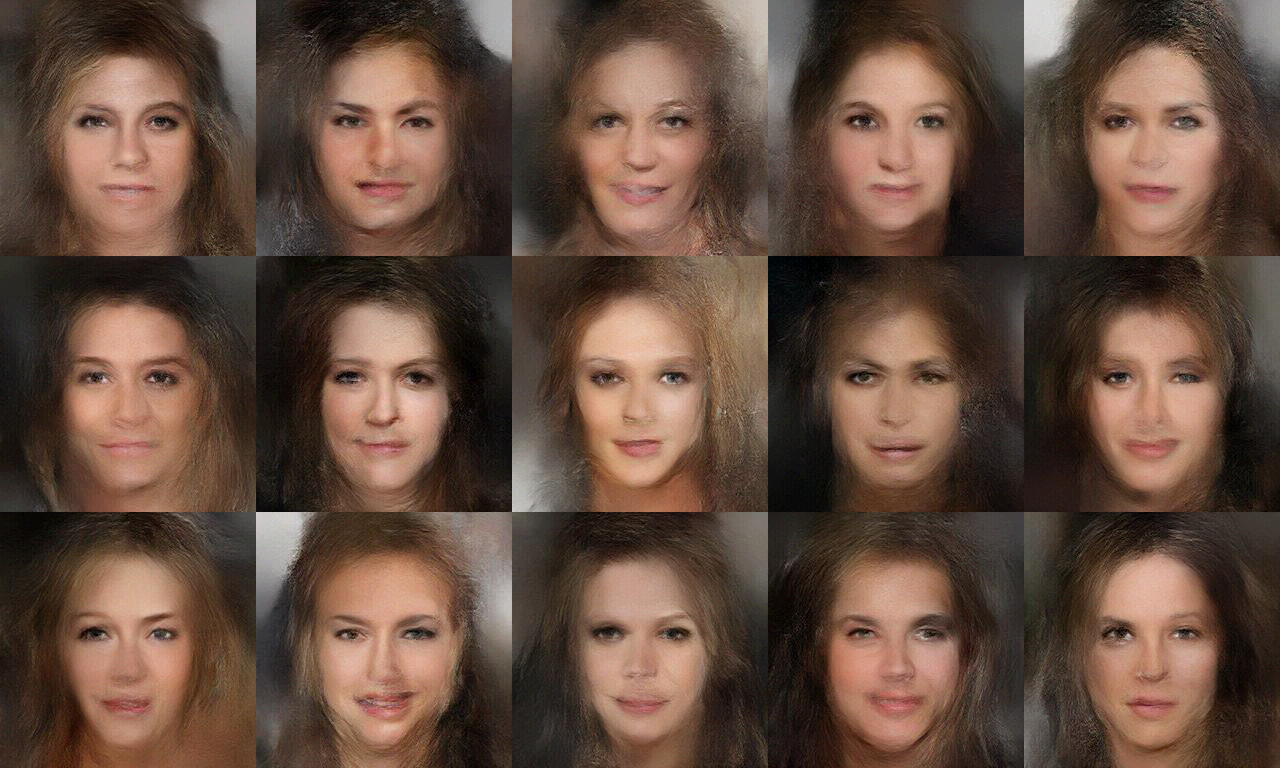}  
		\caption*{}
	\end{subfigure}
	\caption{Selected CelebA-HQ $256 \times 256$ Woodbury-Glow samples (temperature $0.5$).}
	\label{fig:celeba256-sample}
\end{figure*}

In this section, we include additional samples from Woodbury-Glow models trained on our various datasets. These samples complement our quantitative analysis.
We train our models on CIFAR-10~\cite{krizhevsky2009learning}, ImageNet~\cite{russakovsky2015imagenet}, the LSUN church dataset~\cite{yu15lsun}, and the CelebA-HQ dataset~\cite{karras2017progressive}. Specifically, for ImageNet, we use $32 \times 32$ and $64 \times 64$ images. For the LSUN dataset, we use the same approach as \citet{kingma2018glow} to resize the images to be $96 \times 96$. For the CelebA-HQ dataset, we use $64 \times 64$, $128 \times 128$, and $256 \times 256$ images. For LSUN and CelebA-HQ datasets, we use $5$-bit images. The parameter settings of our models are in Table~\ref{tab:hyper} and Table~\ref{tab:w-latent}. The samples are in Figures~\ref{fig:cifar-sample}, \ref{fig:imagenet32-sample}, \ref{fig:imagenet64-sample}, \ref{fig:lsun-sample}, \ref{fig:celeba64-sample}, \ref{fig:celeba128-sample}, and \ref{fig:celeba256-sample}.

\vspace{1cm}

\clearpage

\clearpage

\FloatBarrier

\end{document}